\documentclass[10pt,twocolumn,letterpaper]{article}

\usepackage[pagenumbers]{cvpr} % To force page numbers, e.g. for an arXiv version

\usepackage[dvipsnames]{xcolor}

\definecolor{cvprblue}{rgb}{0.21,0.49,0.74}
\usepackage[pagebackref,breaklinks,colorlinks,citecolor=cvprblue]{hyperref}
\usepackage{amsmath}
\usepackage{float}
\usepackage{stfloats}
\usepackage{afterpage}
\usepackage{array}
\usepackage{lipsum}

\usepackage{soul}

\usepackage{algorithm}
\usepackage{algpseudocode}

\def\paperID{5495} % *** Enter the Paper ID here
\def\confName{CVPR}
\def\confYear{2024}

\newcommand{\cuthalfcaptionup}{\vspace*{-5pt}}
\newcommand{\ignore}[1]{}

\title{Multi-Scale 3D Gaussian Splatting for Anti-Aliased Rendering}

\author{Zhiwen Yan \quad
Weng Fei Low \quad
Yu Chen \quad
Gim Hee Lee\\
Department of Computer Science, National University of Singapore\\
{\tt\small yan.zhiwen@u.nus.edu} \quad {\tt\small \{wengfei.low, chenyu\}@comp.nus.edu.sg} \quad {\tt\small gimhee.lee@nus.edu.sg}
}

\begin{document}
% \maketitle
\twocolumn[{%
\renewcommand\twocolumn[1][]{#1}%
\maketitle
\vspace{-18pt}
% \begin{figure*}[b]
\begin{center}
    \captionsetup{type=figure}
    \cuthalfcaptionup
    \centering
    \includegraphics[width=\textwidth]{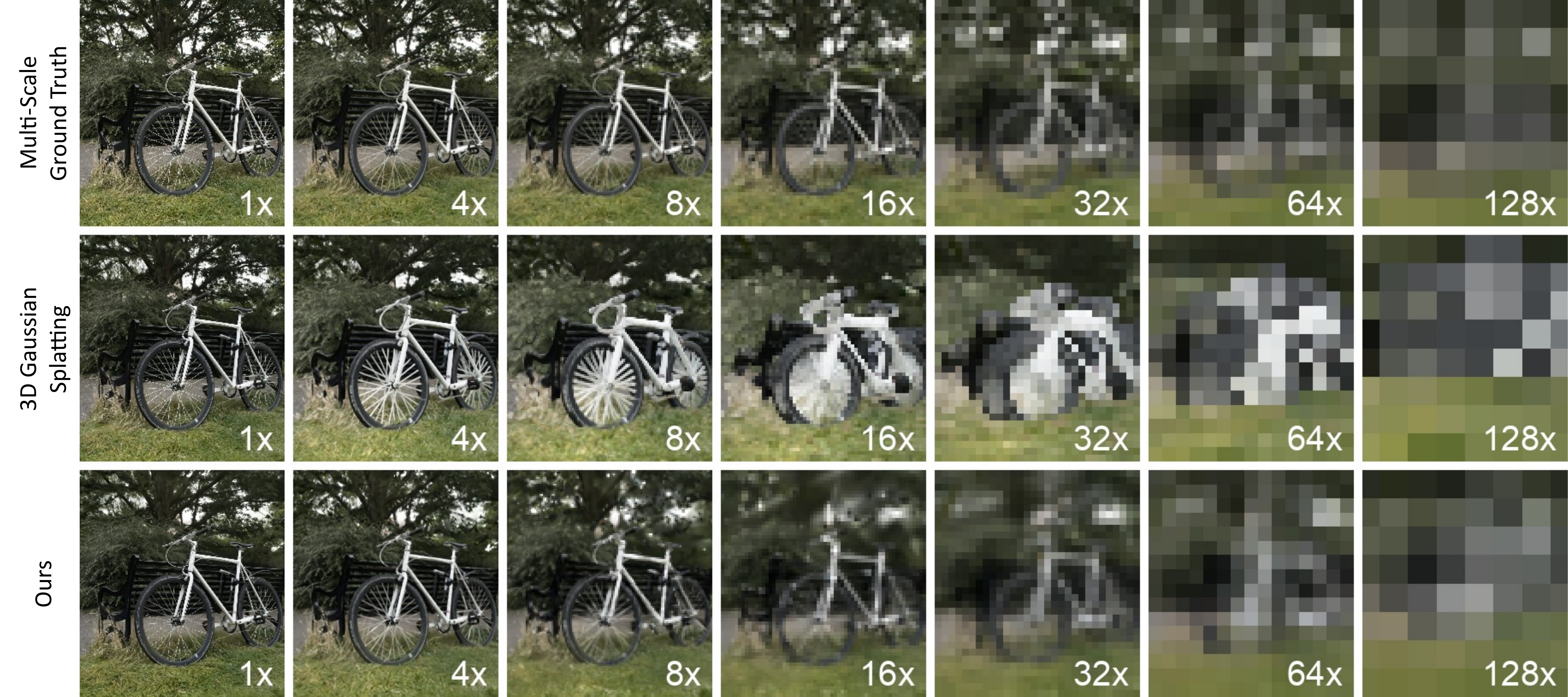}
    % \caption{{\color{red}label yet to be added, need a reference from introduction}}
    \caption{The rendering quality and speed of the original 3D Gaussian splatting\cite{kerbl3Dgaussians} deteriorate severely at low resolutions or from distant cameras due to aliasing. Conversely, our multi-scale 3D Gaussians representation utilizes selective rendering to achieve faster ($160\%-2400\%$ at 4$\times$-128$\times$ resolution) and more accurate rendering at lower resolutions.}
    \cuthalfcaptionup
    \label{fig:teaser}
\end{center}
% \end{figure*}
}]

% \renewcommand{\topfraction}{1.0}
% \renewcommand{\textfraction}{0.1}
% \renewcommand{\floatpagefraction}{1.0}
% \begin{figure*}[b]
%     \centering
%     \includegraphics[width=\textwidth]{images/teaser.pdf}
%     % \caption{{\color{red}label yet to be added, need a reference from introduction}}
%     \caption{The rendering quality and speed of the original 3D Gaussian splatting\cite{kerbl3Dgaussians} deteriorate severely at low resolutions or from distant cameras due to aliasing. Conversely, our multi-scale 3D Gaussians representation utilizes selective rendering to achieve fast and accurate rendering across all resolutions.}
%     \label{fig:teaser}
% \end{figure*}

\begin{abstract}
3D Gaussians have recently emerged as a highly efficient representation for 3D reconstruction and rendering. Despite its high rendering quality and speed at high resolutions, they both deteriorate drastically when rendered at lower resolutions or from far away camera position. During low resolution or far away rendering, the pixel size of the image can fall below the Nyquist frequency compared to the screen size of each splatted 3D Gaussian and leads to aliasing effect. The rendering is also drastically slowed down by the sequential alpha blending of more splatted Gaussians per pixel. To address these issues, we propose a multi-scale 3D Gaussian splatting algorithm, which maintains Gaussians at different scales to represent the same scene. Higher-resolution images are rendered with more small Gaussians, and lower-resolution images are rendered with fewer larger Gaussians. With similar training time, our algorithm can achieve 13\%-66\% PSNR and 160\%-2400\% rendering speed improvement at 4$\times$-128$\times$ scale rendering on Mip-NeRF360 dataset compared to the single scale 3D Gaussian splatting. Project website: \url{https://jokeryan.github.io/projects/ms-gs/}.

\end{abstract}   
\section{Introduction}
\label{sec:intro}

3D Gaussian Splatting\cite{kerbl3Dgaussians} has recently emerged as a highly efficient representation for novel view synthesis. Compared to the time-consuming ray marching used in most neural radiance fields (NeRF)~\cite{mildenhall2020nerf, mueller2022instant, barron2021mipnerf}, a high-resolution image can be rendered in real-time by rasterizing the splatted 3D Gaussians. However, this rasterization algorithm is subjected to severe aliasing effect and speed deterioration when rendering the same scene at low resolution or from distant positions as shown in \cref{fig:teaser}. This limitation significantly constrain the application of the 3D Gaussian splatting algorithm in reconstructing and rendering large-scale scenes. 

Aliasing effect is a consequence of inadequate sampling frequency failing to capture the continuous signal accurately. In the context of rendering, image pixels are sampled with an interval of one-pixel size. The signal can be considered as the 3D scene represented implicitly as in NeRF or explicitly as in 3D Gaussians. When part of the 3D scene is represented with high details but rendered with low resolution or from distant positions, the disparity between the low sampling and high signal frequencies culminates in aliasing artifacts. A naive solution is to render at high resolution and subsequently down-scale the rendered image to a lower resolution. However, this solution is not viable for scenes containing both near and far regions which are very common.
Due to the inability of 3D Gaussian splatting algorithm to accommodate varying resolutions within a single image, rendering the entire image with a even higher resolution for the sake of far away regions is neither time nor memory efficient. 

We postulate that the pronounced aliasing artifacts observed when rendering with 3D Gaussians, as opposed to other techniques such as NeRF, are primarily attributable to the splatting of small Gaussians. 3D regions with intricate details are represented with large amount of small Gaussians. When rendering these regions with low resolution or from a distant view, many splatted small Gaussians are cramped in one pixel and therefore the pixel color of this region is dominated by the front-most Gaussian, even if this Gaussian is much smaller than others and not at the center. This problem is further aggravated by the low pass filter in \cite{kerbl3Dgaussians, zwicker2001ewa} applied to each individual Gaussian with the intention to mitigate aliasing on edges at high resolutions. This problem is explained in more detail in~\cref{sec:aliasing_cause}. 
% On the other hand, NeRF samples many points along the ray or even along the cone volume \cite{barron2021mipnerf, barron2023zipnerf} and is less likely to be affected by the aliasing from the fine details at low resolution. 

In addition to the aliasing artifacts, the rendering speed of 3D Gaussians is also affected at low resolution. The number of 3D Gaussians that need to be rendered remains constant at lower resolutions, but they are more concentrated to fewer pixels. The Gaussians that are splatted to the same pixel cannot be rendered in parallel.
This means that the image rendering is even slower at lower resolution in comparison with NeRF rendering time that reduces linearly with decreasing resolution.
Hence, although aliasing is not a problem exclusive to 3D Gaussian splatting, it is more prominent and more difficult to tackle. 

\begin{figure*}[!ht]
    \centering
    \includegraphics[width=0.9\textwidth]{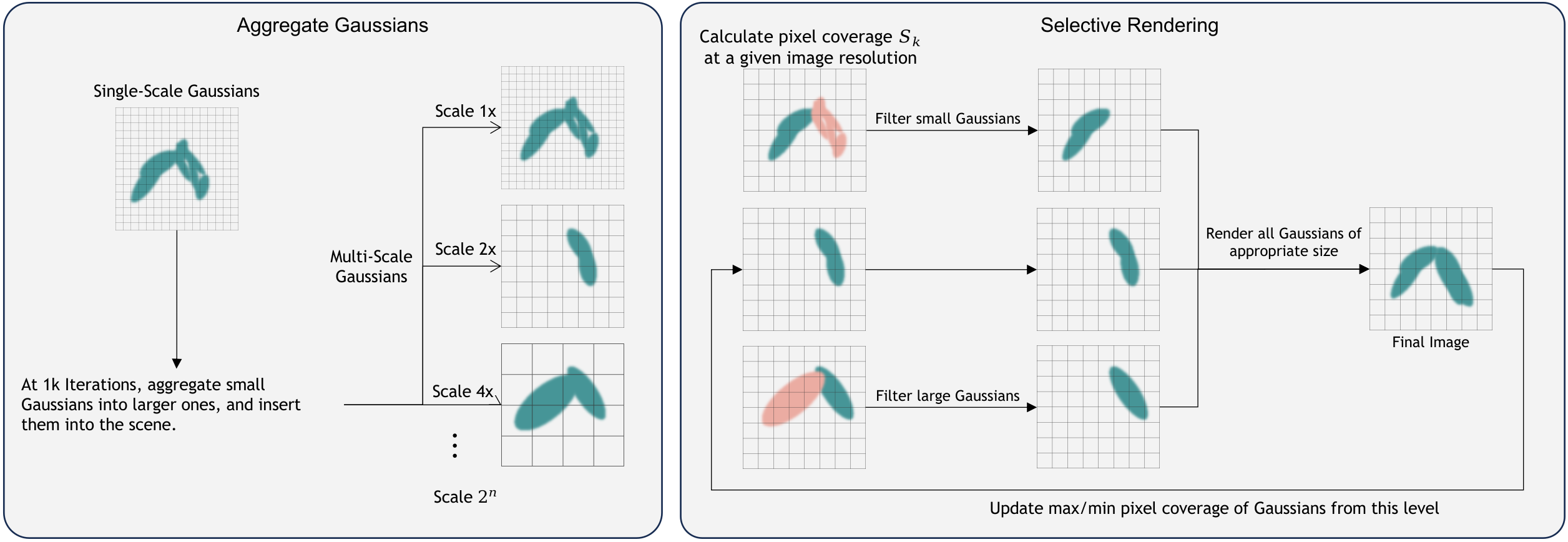}
    \caption{Overall pipeline of our algorithm. At the early stage of training (left), small Gaussians below certain size threshold in each voxel are aggregated, enlarged and inserted into the scene at different resolution scale. During rendering (right), the multi-scale Gaussians of the appropriate ``pixel coverage" at the current render resolution are selected for rendering. If the rendering resolution scale equals to the scale of the Gaussians, the expected ``pixel coverage" range of the Gaussians are updated accordingly.}
    \label{fig:Archi}
    \vspace{-3mm}
\end{figure*}

\paragraph{Contributions}
To mitigate the aliasing problem for 3D Gaussian splatting, we propose a novel multi-scale 3D Gaussians to represent the scene at different levels of detail (LOD) as shown in \cref{fig:Archi}. This is inspired by the mipmap and LOD algorithms widely used in computer graphics, which pre-computes textures and polygons at different scales to be rendered under different resolutions and distances. Similarly, we add larger, coarser Gaussians for lower resolutions by aggregating the smaller and finer Gaussians from higher resolutions. Depending on the pixel coverage of the splatted Gaussians during rendering, only a subset of the Gaussians is used. A simplified explanation for this is that the coarse Gaussians are used to render low-resolution images and the fine Gaussians are used to render high-resolution images. With fewer than 5\% number of Gaussians added and a similar training time, our method can achieve 13\%-66\% PSNR and 160\%-2400\% rendering speed improvements at 4$\times$-128$\times$ scale rendering on Mip-NeRF360 dataset\cite{barron2021mipnerf}, while maintaining a comparable rendering quality and speed at 1$\times$ scale.

\section{Related Works}
\subsection{Anti-Aliasing in Computer Graphics}
Aliasing is a long-standing problem for computer graphics when rendering a scene to a discrete image. Traditional anti-aliasing techniques primarily target mesh representations. Supersampling Anti-Aliasing (SSAA) \cite{Beets2000SupersamplingAA} renders the scene at a higher resolution before downscaling, leading to significantly more time and memory demain, and therefore is less used in real-time applications. The Multisample Anti-Aliasing (MSAA) \cite{realityengine1993msaa, Beets2000SupersamplingAA} algorithm selectively supersamples pixels on the edges, reducing resource and time consumption.
This technique is not very suitable for 3D Gaussian splatting because of its requirement for regular grids and lack of support for variable sampling resolution at different pixels. The more recent Fast Approximate Anti-Aliasing (FXAA) \cite{lottes2011fxaa, jimenez2011filteringapproach} is a post processing algorithm that smooths the jagged edges after the image is rendered. Unfortunately, this technique is also not suitable for Gaussian representation as the front-most Gaussian dominates the pixel color and produces chunky instead of jagged artifacts in mesh rendering. 

In contrast to the supersampling methods mentioned above, our method takes the inspiration from hierarchical mipmap \cite{williams1983mipmap} and level of details (LOD) \cite{tan2004lod, erikson1996polygon} algorithms to address the aliasing for 3D Gaussians. Mipmap uses multi-scale textures for the rendering at different resolution or from different distances. LOD algorithm represents the models in a scene with different complexity to be rendered at different distances. Both techniques not only mitigate the aliasing effect by reducing the complexity of the scene representation, but also enhances rendering speed, particularly for large-scale scenes. 

\subsection{Anti-Aliasing in Neural Representation}
The recent success of neural representations especially Neural Radiance Fields (NeRF) \cite{mildenhall2020nerf, mueller2022instant, Chen2022tensorrf} has also inspired some works to develop algorithms against aliasing effect on neural representations beyond the traditional mesh representation. Mip-NeRF \cite{barron2021mipnerf,barron2022mipnerf360} employ low pass filters on the positional encoding of the input spatial coordinates to reduce the scene signal frequency. Building on the hash grid representation used by InstantNGP \cite{mueller2022instant} with no position encoding, Zip-NeRF \cite{barron2023zipnerf} proposes a multi-sampling strategy in the conical frustum instead of the camera ray, at the cost of 6$\times$ rendering time. Similar to the mipmap algorithm in mesh texture rendering, Tri-MipNeRF \cite{hu2023trimipnerf} and MipGrid \cite{nam2023mipgrid} proposes to use multi-scale feature grids for rendering at different resolution or distance. 

Conversely, 3D Gaussian splatting \cite{kerbl3Dgaussians} presents unique anti-aliasing challenges due to its distinct scene representation. It does not have any positional encoding or feature grid, and its requirement for regular grids conflicts the more flexible multi-sampling strategies. 
The concentration of small Gaussians in detail-rich regions exacerbates aliasing and speed issues, more so than in NeRF representations. 
To the best of our knowledge, we are the first to propose an anti-aliasing algorithm for scene reconstruction using 3D Gaussian splatting.

\section{Preliminaries}
\label{sec:preliminaries}
\subsection{3D Gaussian Splatting}
3D Gaussian splatting is first proposed in EWA Splatting \cite{zwicker2001ewa}, and
later used by \cite{kerbl3Dgaussians} for scene reconstruction and novel view synthesis. The scene is represented by a set of $\mathbf{K}$ 3D Gaussians $\{ \mathcal{G}_{\hat{V}^k, \hat{\mu}^k}, \sigma_k, \mathbf{c}_k  \mid  k \in [1,\mathbf{K}] \}$ with variance $\hat{V}^k$, center $\hat{\mu}^k$, density $\sigma_k$ and color $\mathbf{c}^k$. During rendering, the 3D Gaussians are splatted to the 2D screen to by the perspective transformation to form 2D Gaussians $\mathcal{G}_{V^k, \mu^k}$. The image is then divided into 16$\times$16 regular tiles and all 2D Gaussians touching each tile are sorted based on their original depth. The color of each pixel in the tile is then rasterized from the sequential alpha blending the 2D Gaussians from front to back.

\subsection{Cause of Aliasing in 3D Gaussian Splatting}
\label{sec:aliasing_cause}
Aliasing can occur when sampling a continuous signal $g(x)$ with a discrete sampling function $\delta_s(x, \Delta x)=\sum_{n=-\infty}^{\infty}\delta(x-n\cdot \Delta x)$, where $\delta$ is a impulse function. The result of the sample in the spatial domain is:
\begin{equation}
    g_s(x)=\delta_s(x, \Delta x) \cdot g(x) .
\end{equation}
This sampled function converted into the frequency domain using Fourier transform operator $\mathcal{F}$ becomes:
\begin{equation}
    \begin{aligned}
        \mathcal{F}[{g_s}(u)] &= \frac{1}{\Delta x} \sum_{k=-\infty}^{\infty}{\delta(u-\frac{k}{\Delta x})} \ast \mathcal{F}[g(x)] \\
                              &= \frac{1}{\Delta x} \sum_{k=-\infty}^{\infty}\mathbf{G}(u-\frac{k}{\Delta x}).
    \end{aligned}
\end{equation}
When the highest frequency component $f_{max}$ of the signal is greater than half of the sampling frequency $f_s=\frac{1}{\Delta x}$, $\mathbf{G}(u-\frac{k}{\Delta x})$ in the summation sequence would overlap with each other and causes the sampled signal to diverge from the actual signal. This phenomenon is the aliasing effect and the minimum sampling frequency needed to avoid aliasing is $f_{Ny}=2\cdot f_{max}$, known as the Nyquist frequency.

% Based on the above discussion, the two most common approaches to mitigate the aliasing effect are to increase the sampling frequency or to reduce the signal frequency. Zip-NeRF\cite{barron2023zipnerf} increases the sampling frequency by multi-sampling, but the signal representing the 3D scene is not band-limited in most cases and often does not have the highest frequency component. The super-sampling strategy is not only time-consuming but also not guaranteed to achieve the Nyquist frequency. Mip-NeRF\cite{barron2021mipnerf} choose to reduce the signal frequency by applying a low pass filter to the positional encoding, but this approach cannot be easily applied to 3D Gaussians. 3D Gaussians are explicit discrete representations where each Gaussian can contain both high-frequency and part of the low frequency signal. We will demonstrate in \cref{sec:exp} that naively filtering out the high-frequency Gaussians will cause the low-frequency signal to be lost as well.

The EWA splatting \cite{zwicker2001ewa} used by 3D Gaussian splatting \cite{kerbl3Dgaussians} also tries to mitigate the aliasing problem by applying a low pass filter to the rendered color. 
To approximate this efficiently, it applies a Gaussian kernel $h(\mathbf{x})$ as the low pass filter on each splatted 2D signal $g_c(\mathbf{x})$ independently to produce a band limited signal: %$g'_c(\mathbf{x})$:
\begin{equation}
    \begin{aligned}
        g_c'(\mathbf{x}) &= g_c(\mathbf{x}) \ast h(\mathbf{x}) \\
                         % &= \int_{\mathcal{R}^2}{g_c(\eta)\cdot h(\mathbf{x}-\eta)}d\eta, \\
                         &\approx \sum_{k}\sigma_k \mathbf{c}_k T_k \int_{\mathcal{R}^2}q_k(\eta)h(\mathbf{x}-\eta)d\eta \\
                         &= \sum_{k}\sigma_k \mathbf{c}_k T_k \cdot (q_k \ast h)(\mathbf{x}),
    \end{aligned}
\end{equation}
where $\mathcal{R}^2$ is the range of one pixel, $q_k(\mathbf{x})$ is the 2D integrated Gaussian kernel, and $\sigma_k, \mathbf{c}_k, T_k$ are the opacity, color, and transmittance at each Gaussian, respectively. By combining the reconstruction Gaussian kernel $\mathcal{G}_{V^k}$ and low pass Gaussian kernel $\mathcal{G}_{V^h}$ of covariance matrix $V^k$ and $V^h$, the band limit function becomes:
\begin{equation}
    \begin{aligned}
        g_c'(\mathbf{x}) &= \sum_k \alpha_k \cdot (\mathcal{G}_{V^k} \ast \mathcal{G}_{V^h})(\mathbf{x}) \\ 
                         &= \sum_k \alpha_k \cdot \mathcal{G}_{V^k+V^h}(\mathbf{x}),
    \end{aligned}
\end{equation}
where $\alpha_k$ represents all coefficients invariant of $\mathbf{x}$ at each Gaussian and $V^h$ is determined by the screen pixel size. A simple understanding of this is that the covariance of each 3D Gaussian is increased based on the screen pixel size. 

This method of applying a low pass filter to each 3D Gaussian independently helps to smooth the edges of the Gaussians when the Gaussians are not too small compared to the pixel size. However, it also gives rise to two substantial issues at low resolutions: 
\begin{enumerate}
    \item %The 
    $V^h$ added to the original covariance $V^k$ effectively increases the extent of each Gaussian, especially when $V^h$ is large compared to $V^k$ at low resolutions. Small Gaussians in the front %will 
    dominate the color of the pixel and cause severe artifacts shown in \cref{fig:360_qualitative}.
    % {\color{red}figure showing artifacts}
    % \item the total number of Gaussian remains unchanged but more Gaussians in each pixel must be summed with $\sum_k$ sequentially at low resolution because of the incremental calculation of the transmittance $T_k$. This makes the rendering at low resolution even slower than at high resolution. 
    \item The number of Gaussians involved in the sequential $\sum_k$ for each pixel scales increases with decreasing image resolution. 
    Due to the incremental calculation of the transmittance $T_k$, the rendering even slower at lower resolutions.
    % {\color{red}figure of speed at different resolution}
\end{enumerate}

\section{Our Method}
\label{sec:method}
\subsection{Multi-Scale Gaussians Based on Pixel Coverage}
To mitigate the aliasing artifacts of 3D Gaussians \cite{kerbl3Dgaussians} while avoiding the two problems of the EWA splatting \cite{zwicker2001ewa}, we introduce multi-scale 3D Gaussians (\cf \cref{fig:Archi}) that tackle the problem on the scene-level instead of on each individual Gaussian. The 3D scene is represented with Gaussians from 4 levels of detail, corresponding to the $1\times$, $4\times$, $16\times$, and $64\times$ downsampled resolution. Small finer-level Gaussians are aggregated to create larger Gaussians for coarser levels during training.  Each 3D Gaussian $\mathcal{G}_{k}^{l}$ belongs to one of the levels $l$ and is included or excluded independently during the rendering based on its ``pixel coverage". 

\begin{figure}
    \centering
    \includegraphics[width=\linewidth]{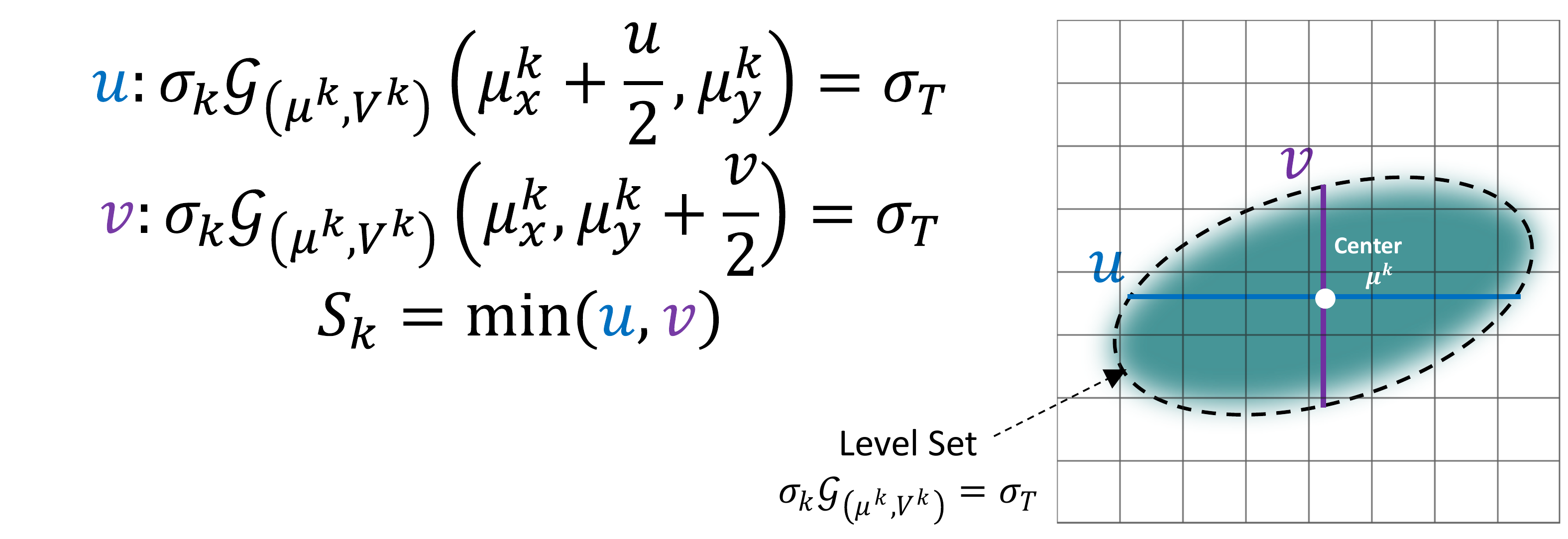}
    \caption{Pixel coverage of a 3D Gaussian is its horizontal or vertical size, whichever is smaller measured by the level set.}
    \label{fig:pixel_coverage}
\end{figure}

\vspace{-3mm}
\paragraph{Pixel Coverage of Gaussian.}
The ``pixel coverage" of a Gaussian reflects the size of the Gaussian when splatted onto the screen space compared to the pixel size at the current rendering resolution. The ``pixel coverage" $S_k$ of a splatted 2D Gaussian $\mathcal{G}_{(\mu^k, V^k)}$ is defined as the length of its horizontal or vertical axis until the low opacity level set, whichever is smaller, as shown in \cref{fig:pixel_coverage}.
The pixel coverage is measured in pixel count and the opacity threshold $\sigma_T$ is set as $\frac{1}{255}$. 
% :
% \begin{equation}
%     \begin{aligned}
%     S_k = min(&\{u|u>0, \sigma_k \mathcal{G}_{(\mu^k, V^k)}(\mu^k_x + u/2, \mu^k_y) = \sigma_{T}\}, \\
%               &\{v|v>0, \sigma_k \mathcal{G}_{(\mu^k, V^k)}(\mu^k_x, \mu^k_y + v/2) = \sigma_{T}\})  \\
%         = min(&\sqrt{-\frac{2}{V_{11}^k}log(\frac{\sigma_{T}}{\sigma_k})}, 
%         \sqrt{-\frac{2}{V_{22}^k}log(\frac{\sigma_{T}}{\sigma_k})}),
%     \end{aligned}
% \end{equation}

% The minimum between the $x$ and $y$ axis direction coverage is chosen because the highest frequency of depends on the smallest spatial span.

The pixel coverage approximates the extent of a 2D splatted Gaussian in the spatial domain. During the rendering from a given camera direction, the color of each splatted Gaussian is constant within this pixel coverage. %Hence, this pixel coverage
As a result, the coverage of this pixel approximates the inverse of the highest frequency component $f_{max}=1/S_k$ in this region. Compared to the sampling frequency of $f_s = 1\mathrm{px}^{-1}$ during rasterization, a signal frequency of $f_{max} > f_s/2$ can cause the sampling to fall below the Nyquist frequency needed to avoid aliasing.

\begin{figure}
    \centering
    \includegraphics[width=0.9\linewidth]{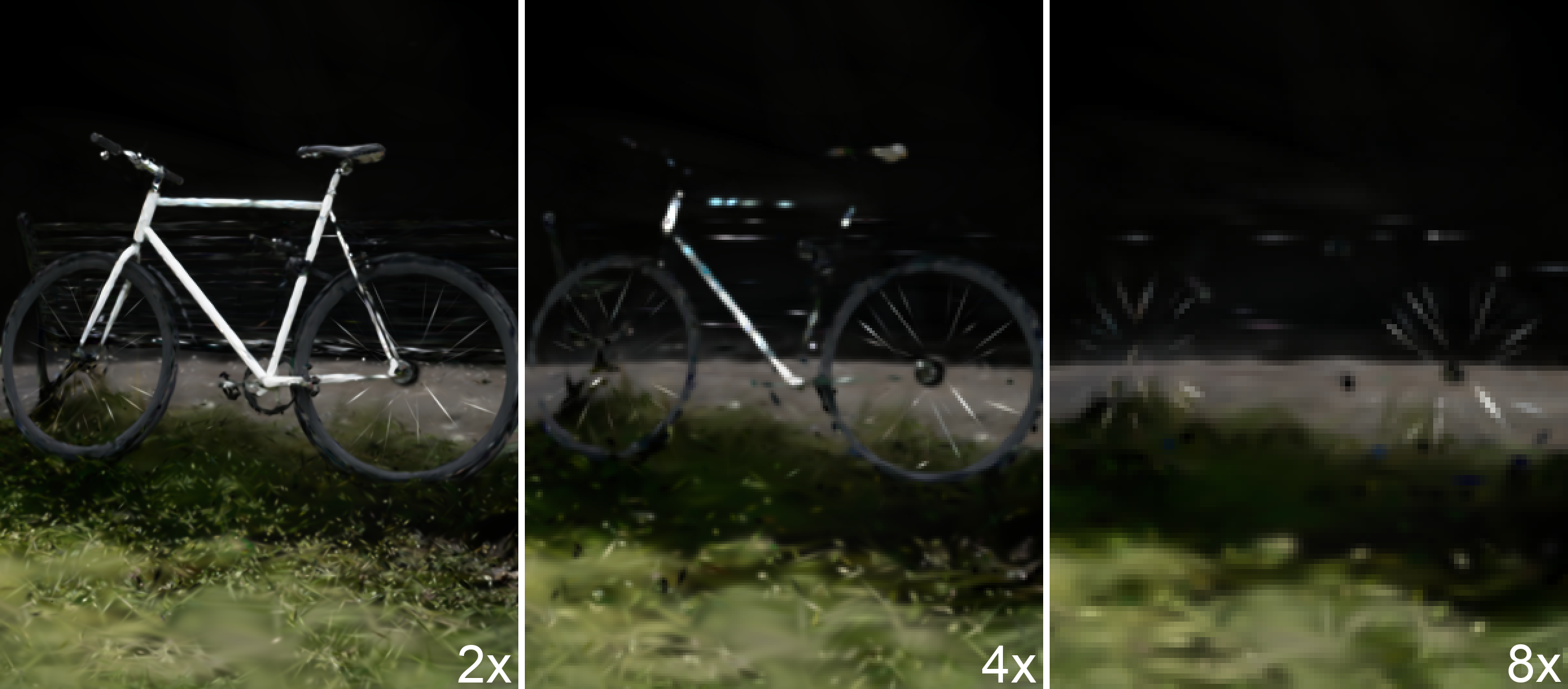}
    \caption{
    Missing parts caused by naive small Gaussian filtering at different resolution scales.}
    \label{fig:dis_bicycle}
    \vspace{-2mm}
\end{figure}

Consequently, the Gaussians with pixel coverage $S_k < S_T = 2\mathrm{px}$ should be filtered out during rendering to avoid aliasing. Since 3D Gaussian representation does not encode the signal of different frequencies at different Gaussians, naively filtering out the small Gaussians will result in a hole or part missing in the scene as shown in \cref{fig:dis_bicycle}. To address this issue, we propose to aggregate the small Gaussians to form large Gaussians that encode the low-frequency signal. These large Gaussians would appear when the small Gaussians are filtered out. 

\begin{figure}
    \centering
    \includegraphics[width=\linewidth]{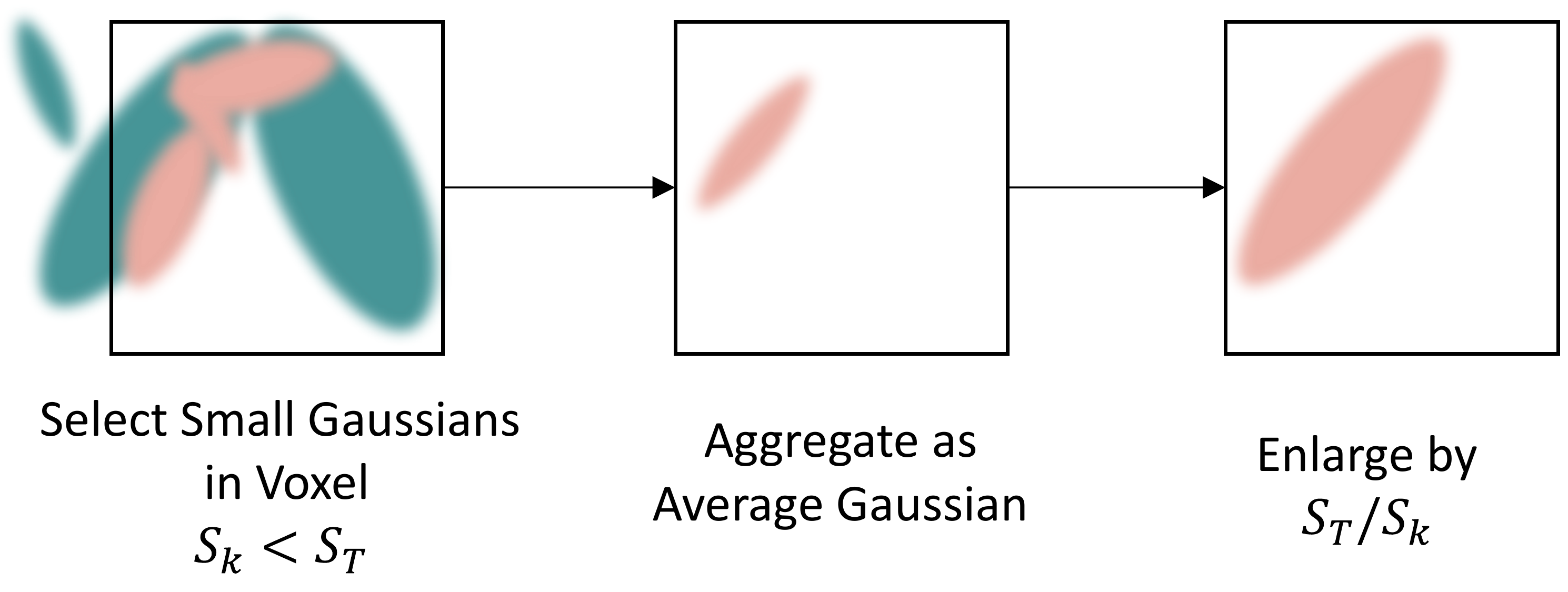}
    \caption{Large Gaussians are created by aggregating the small Gaussians in each voxel below the pixel coverage threshold, and then enlarged by the pixel coverage multiplier.}
    \label{fig:aggregate}
\end{figure}

\vspace{-3mm}
\paragraph{Aggregate to Insert Large Gaussians.}
All 3D Gaussians initialized from the input point cloud at the start of the training belong to the finest level $l=1$. They are densified by splitting and cloning as in \cite{kerbl3Dgaussians}, and all the densified Gaussians would inherit the same level. After the warm-up stage of the first 1,000 iterations, we introduce coarse-level Gaussians by aggregating fine-level Gaussians that are too small as visualized in \cref{fig:aggregate} and described in \cref{alg:aggregate}. The procedure is outlined as follows:
\begin{enumerate}
    \item For all levels $\{l_m \mid 2 \le l_m \le l_{max}\}$, we render all 3D Gaussians from $[1,l_m-1]$ at the $4^{l_m-1}$ times downsampled resolution of all training images. All 3D Gaussians with the minimal ``pixel coverage" $S_k$ smaller than the filter threshold $S_T$ are chosen for the aggregation. 
    % \item The chosen Gaussian positions $x_k$ are normalized from $[-\infty, \infty]$ to $[-2, 2]$, by linearly mapping $[-x_{scene}, x_{scene}]$ to $[-1, 1]$ and non-linearly mapping the rest using $x'_k=2-\frac{x_{scene}}{x_k}$, where $x_{scene}$ is the scene extent determined by camera position range. 
    % \item Voxelize the normalized chosen Gaussians with a voxel resolution $0.005\cdot l_m$. The attributes of all Gaussians within each voxel are aggregated using average pooling.
    \item The chosen 3D Gaussians are binned by a $(400/l_m)^3$ resolution voxel grid based on their positions. The attributes of all Gaussians within each voxel are aggregated to create a new Gaussian using average pooling, including position, scaling, opacity and color. More details are included in the supplementary.
    \item Based on the average ``pixel coverage" $S_{avg}$ of Gaussians in each voxel, the scaling of each new Gaussian created is enlarged by $S_T/S_{avg}$ so that it is of a size suitable to be rendered at $l_m$. This new Gaussian belongs to level $l_m$. 
\end{enumerate}
Not all Gaussians from the fine levels are small. Many Gaussians in the background or in the textureless regions are large and do not need to be aggregated. The number of Gaussians created is often fewer than 5\% of the final total number of Gaussians.

\begin{algorithm}
\caption{Aggregate Small Gaussians}
\label{alg:aggregate}
\begin{algorithmic}[1]
\Procedure{AggregateGaussians}{$\mathcal{G}_{1:K}^{1:l_{max}}$}
    % \State $z \gets x + y$
    \For{$l_m \gets 2$ \textbf{to} $l_{max}$}
        \State $S_{1:K}$=PixelCoverage($\mathcal{G}_{1:K}^{1:l_m-1}$, scale $4^{l_m-1}$)
        \State $G_{small}=\{\mathcal{G}_k|S_k<S_T, \forall k \in [1:K]\}$
        \For{$n \gets 1$ \textbf{to} $(400/l_m)^3$}
            \State $G_{n}=\{ \mathcal{G}_k | \mathcal{G}_k \,$in voxel$\, n, \forall \mathcal{G}_k \in G_{small}\}$
            \State $\mathcal{G}_{new,n}^{l_m}=$ Enlarge(Average($G_{n}$))
            \State InsertIntoScene($\mathcal{G}_{new,n}^{l_m}$)
        \EndFor
    \EndFor
\EndProcedure
\end{algorithmic}
\end{algorithm}

\begin{algorithm}
\caption{Selective Rendering Based on Pixel Coverage}
\label{alg:selective}
\begin{algorithmic}[1]
\Procedure{SelectiveRender}{$\mathcal{G}_{1:K}^{1:l_{max}}$, scale $l_{r}$}
    \State $S_{1:K}=$PixelCoverage$(\mathcal{G}_{1:K}^{1:l_{max}})$
    \State $G_{1}=\{ \mathcal{G}_k | S_k / S_k^{max} \le S_{rel}^{max}, \forall k \}$
    \State $G_{2}=\{ \mathcal{G}_k | S_k / S_k^{min} \ge S_{rel}^{min} \lor S_k \ge S_T, \forall k \}$
    \State $G_{l_r}=\{ \mathcal{G}_k^l | l=l_r, \forall k \}$
    \For{$\mathcal{G}_k^l \in G_{l_r}$}
        \State UpdateRange($S_k^{max}, S_k^{min}, S_k$)
    \EndFor
    \State \textbf{return} Render($G_1 \cap G_2, l_r$)
\EndProcedure
\end{algorithmic}
\end{algorithm}

\vspace{-3mm}
\paragraph{Multi-Scale Training and Selective Rendering.}
After the large Gaussians are added, the model is trained with both the original images and the downsampled images. A maximum pixel coverage $S_k^{max}$ and a minimum pixel coverage $S_k^{min}$ of each Gaussian are stored for the selective rendering. If the rendering downsample scale equals to the downsample scale when the Gaussian $\mathcal{G}_k$ is created, its $S_k^{max}$ and $S_k^{min}$ values are updated with the new pixel coverage $S_k$:
\begin{equation}
    \begin{aligned}
        S_k^{'max} &= max(\lambda_1 S_k^{max}, S_k), \\
        S_k^{'min} &= min(\lambda_2 S_k^{min}, S_k),
    \end{aligned}
\end{equation}
where $\lambda_1$ and $\lambda_2$ are decay coefficients taking the empirical value of $0.95$ and $1.05$, respectively. 

\begin{figure}
    \centering
    \includegraphics[width=\linewidth]{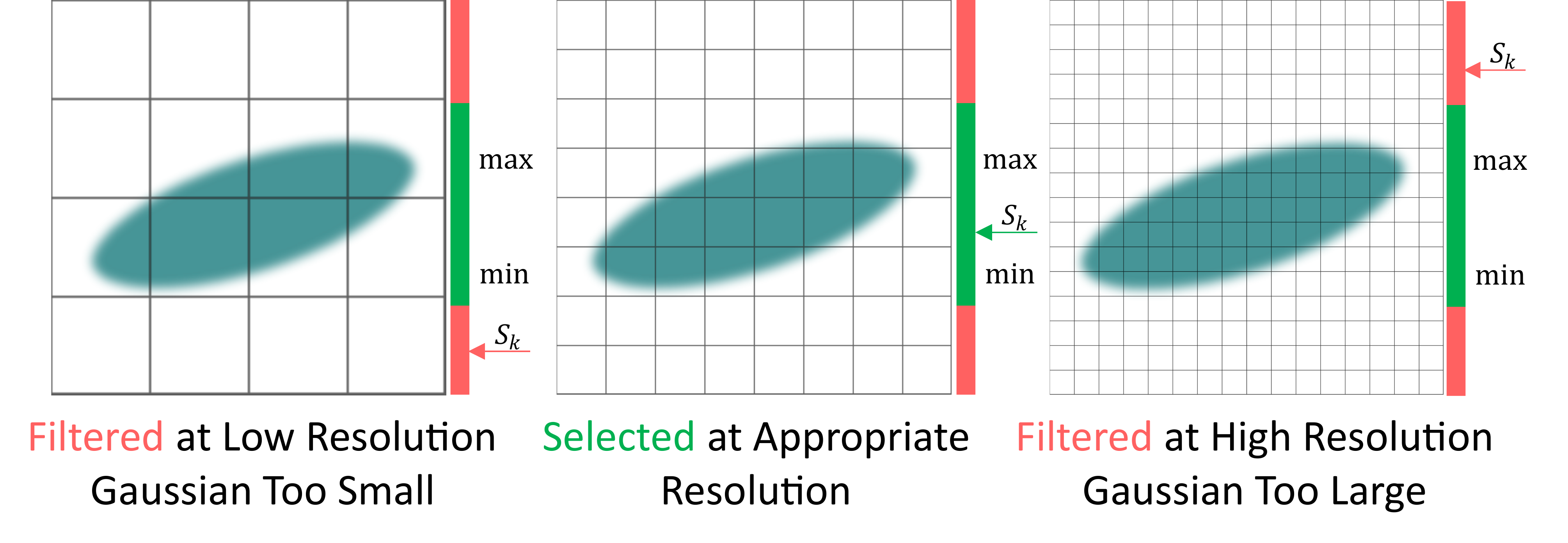}
    \caption{Based on the rendering resolution, the current pixel coverage of a Gaussian relative to its minimum and maximum pixel coverages determines whether it is selected for rendering.}
    \label{fig:filter}
    \vspace{-3mm}
\end{figure}

During rendering at any resolution or camera distance, a Gaussian is selected for rendering if its pixel coverage $S_k$ on the screen satisfies the following condition:
\begin{equation}
        (\frac{S_k}{S_k^{max}} \leq S_{rel}^{max}) \land (\frac{S_k}{S_k^{min}} \geq S_{rel}^{min} \lor S_k \geq S_T),
\end{equation}
where $S_{rel}^{max}$ and $S_{rel}^{min}$ are the maximum and minimum relative pixel coverage taking the empirical values of $1.5$ and $0.5$ respectively. If the pixel coverage of a Gaussian is too much larger than the $S_k^{max}$, it is filtered out from rendering. Similarly, if it is too much smaller than the $S_k^{min}$ and is smaller than $S_T$, it is filtered out from rendering (\cf \cref{fig:filter}). The absolute $S_T$ threshold is used to preserve the large Gaussians from the lower scales, as they do not cause the aliasing problem if their screen size is not sufficiently small. This selective rendering procedure is described in \cref{alg:selective}.
% Additionally, to render beyond the maximum and below the minimum training resolution, 
% the $S_k^{max}$ of finest level Gaussians and the $S_k^{min}$ of the coarsest level Gaussians are not updated. This means that
% the Gaussians from the finest level are never filtered even if it is too large, and Gaussians from the coarsest level are never filtered even if it is too small. 
Additionally, even if the Gaussians from the finest level are too large or the Gaussians from the coarsest level are too small, they are not filtered to render beyond the maximum and below the minimum training resolutions.

The pixel coverage range of each Gaussian allows the model to maintain multi-scale Gaussians for different levels of detail. The appropriate subset of Gaussians is chosen for rendering at different resolutions and distances. More smaller Gaussians encoding the high-frequency information are rendered at high resolution, and fewer and larger Gaussians encoding the low-frequency information are rendered at low resolution for less aliasing effect and faster speed.  
\newcolumntype{C}[1]{>{\centering\arraybackslash}p{#1}} % for center alignment
\newcolumntype{R}[1]{>{\raggedleft\arraybackslash}p{#1}} % for right alignment

\begin{table*}[!htp]
\centering
\caption{Quantitative comparison and ablation study on the 360 dataset~\cite{barron2022mipnerf360} at various downsampled scales, with time in ``ms".}
\vspace{-3mm}
\label{tab:mipnerf-comparison}
\begin{tabular}{lR{0.75cm}R{0.75cm}R{0.75cm}R{0.75cm}R{0.75cm}R{0.75cm}R{0.75cm}R{0.75cm}R{0.75cm}R{0.75cm}R{0.75cm}R{0.75cm}}
\toprule
Scale & \multicolumn{3}{c}{1x} & \multicolumn{3}{c}{4x} & \multicolumn{3}{c}{16x} & \multicolumn{3}{c}{64x} \\
\cmidrule(lr){2-4} \cmidrule(lr){5-7} \cmidrule(lr){8-10} \cmidrule(lr){11-13}
Metric & \small{PSNR$\uparrow$} & \small{LPIPS$\downarrow$} & \small{Time$\downarrow$} & \small{PSNR$\uparrow$} & \small{LPIPS$\downarrow$} & \small{Time$\downarrow$} & \small{PSNR$\uparrow$} & \small{LPIPS$\downarrow$} & \small{Time$\downarrow$} & \small{PSNR$\uparrow$} & \small{LPIPS$\downarrow$} & \small{Time$\downarrow$} \\
\midrule
3D Gaussian\cite{kerbl3Dgaussians} & \textbf{27.52} & \textbf{0.142} & 10.5 & 22.50 & 0.137 & 9.3 & 17.79 & 0.149 & 27.9 & 15.23 & N.A. & 103.3 \\
3DGS + MS Train & 27.35 & 0.155 & 11.3 & 23.50 & \textbf{0.126} & 7.7 & 20.21 & 0.115 & 22.8 & 19.38 & N.A. & 84.8 \\
3DGS + Filter Small & 27.40 & 0.153 & 10.0 & 23.81 & 0.149 & 5.4 & 20.02 & 0.186 & 4.8 & 17.38 & N.A. & \textbf{4.6} \\
3DGS + Insert Large & 18.02 & 0.604 & 9.7 & 18.75 & 0.531 & \textbf{2.5} & 20.23 & 0.256 & \textbf{2.7} & 21.53 & N.A. & 7.1 \\
Our Full Method & 27.39 & 0.155 & \textbf{9.1} & \textbf{24.82} & 0.132 & 5.4 & \textbf{24.75} & \textbf{0.066} & 4.9 & \textbf{25.35} & N.A. & 4.9 \\
\bottomrule
\end{tabular}
\end{table*}

\begin{table*}[!htp]
\centering
\caption{Quantitative comparison and ablation study on Tank and Temples dataset~\cite{Knapitsch2017tank} at various downsampled scales, with time in ``ms".}
\vspace{-3mm}
\label{tab:tnt-comparison}
\begin{tabular}{lR{0.75cm}R{0.75cm}R{0.75cm}R{0.75cm}R{0.75cm}R{0.75cm}R{0.75cm}R{0.75cm}R{0.75cm}R{0.75cm}R{0.75cm}R{0.75cm}}
\toprule
Scale & \multicolumn{3}{c}{1x} & \multicolumn{3}{c}{4x} & \multicolumn{3}{c}{16x} & \multicolumn{3}{c}{64x} \\
\cmidrule(lr){2-4} \cmidrule(lr){5-7} \cmidrule(lr){8-10} \cmidrule(lr){11-13}
Metric & \small{PSNR$\uparrow$} & \small{LPIPS$\downarrow$} & \small{Time$\downarrow$} & \small{PSNR$\uparrow$} & \small{LPIPS$\downarrow$} & \small{Time$\downarrow$} & \small{PSNR$\uparrow$} & \small{LPIPS$\downarrow$} & \small{Time$\downarrow$} & \small{PSNR$\uparrow$} & \small{LPIPS$\downarrow$} & \small{Time$\downarrow$} \\
\midrule
3D Gaussian\cite{kerbl3Dgaussians} & 23.74 & \textbf{0.096} & 6.5 & 19.70 & 0.105 & 11.1 & 15.61 & 0.068 & 43.4 & 13.88 & N.A. & 82.6 \\
3DGS + MS Train & 22.97 & 0.118 & 6.0 & 21.46 & \textbf{0.086} & 9.6 & 18.56 & 0.049 & 37.4 & 16.54 & N.A. & 71.7 \\
3DGS + Filter Small & \textbf{23.78} & 0.100 & 5.6 & 20.12 & 0.107 & 4.5 & 17.41 & 0.072 & 4.4 & 14.95 & N.A. & 4.7 \\
3DGS + Insert Large & 10.84 & 0.697 & \textbf{5.1} & 11.15 & 0.703 & \textbf{1.7} & 11.73 & 0.447 & \textbf{1.7} & 12.62 & N.A. & \textbf{2.5} \\
Our Full Method & 23.46 & 0.111 & 7.6 & \textbf{21.92} & 0.087 & 4.7 & \textbf{20.91} & \textbf{0.034} & 4.8 & \textbf{19.67} & N.A. & 5.9 \\
\bottomrule
\end{tabular}
\end{table*}

\begin{table*}[!htp]
\centering
\caption{Quantitative comparison and ablation study on the Deep Blending dataset~\cite{deepblending} at various downsampled scales, with time in ``ms".}
\vspace{-3mm}
\label{tab:db-comparison}
\begin{tabular}{lR{0.75cm}R{0.75cm}R{0.75cm}R{0.75cm}R{0.75cm}R{0.75cm}R{0.75cm}R{0.75cm}R{0.75cm}R{0.75cm}R{0.75cm}R{0.75cm}}
\toprule
Scale & \multicolumn{3}{c}{1x} & \multicolumn{3}{c}{4x} & \multicolumn{3}{c}{16x} & \multicolumn{3}{c}{64x} \\
\cmidrule(lr){2-4} \cmidrule(lr){5-7} \cmidrule(lr){8-10} \cmidrule(lr){11-13}
Metric & \small{PSNR$\uparrow$} & \small{LPIPS$\downarrow$} & \small{Time$\downarrow$} & \small{PSNR$\uparrow$} & \small{LPIPS$\downarrow$} & \small{Time$\downarrow$} & \small{PSNR$\uparrow$} & \small{LPIPS$\downarrow$} & \small{Time$\downarrow$} & \small{PSNR$\uparrow$} & \small{LPIPS$\downarrow$} & \small{Time$\downarrow$} \\
\midrule
3D Gaussian\cite{kerbl3Dgaussians} & 29.65 & \textbf{0.094} & 8.6 & 27.48 & 0.066 & 7.5 & 22.06 & 0.067 & 20.7 & 17.75 & N.A. & 59.7 \\
3DGS + MS Train & 29.46 & 0.102 & 6.6 & 28.18 & \textbf{0.062} & 5.3 & 24.13 & 0.055 & 14.3 & 20.03 & N.A. & 41.3 \\
3DGS + Filter Small & 29.68 & 0.095 & 6.7 & 28.26 & 0.064 & 4.2 & 24.52 & 0.078 & 3.6 & 18.29 & N.A. & \textbf{3.2} \\
3DGS + Insert Large & 20.59 & 0.379 & \textbf{4.6} & 20.83 & 0.336 & \textbf{1.6} & 21.29 & 0.143 & \textbf{2.1} & 20.10 & N.A. & 4.2 \\
Our Full Method & \textbf{29.70} & 0.096 & 7.4 & \textbf{28.43} & 0.064 & 3.9 & \textbf{27.66} & \textbf{0.036} & 3.4 & \textbf{25.70} & N.A. & 3.4 \\
\bottomrule
\end{tabular}
\end{table*}

\section{Experiments}%Results}}
\label{sec:exp}

In this section, we present a comprehensive evaluation of our proposed model, which is grounded on the implementation framework of the official release of the 3D Gaussian Splatting code. To achieve a similar training time as the baseline model, our models are trained for 40000 iterations with all other hyper-parameters unchanged. All rendering speed are measured on a single RTX3090 GPU. 
We evaluate the performance of the vanilla 3D Gaussian Splatting\cite{kerbl3Dgaussians} algorithm and our model on the multi-scale 360\cite{barron2022mipnerf360}, Tank And Temples\cite{Knapitsch2017tank}, and Deep Blending\cite{deepblending} dataset, aligned with the data used by the original paper. These datasets cover a wide range of object centric, indoor, and ourdoor scenes. 

Our evaluation focuses on the rendering quality and speed at multiple downsampling scales of 1x, 4x, 16x, and 64x %, 
derived from the test views. The rendering quality is measured in PSNR and LPIPS, while the speed is measured in per-image rendering time.
This multi-scale evaluation is aimed at simulating the rendering performance in scenarios of low-resolution imaging or when captured from distant cameras. 
More detailed evaluations, including the results for more resolution scales and per-scene decomposition, are included in the supplementary materials due to the space constraint. Additionally, the supplementary materials include a video that offers an intuitive qualitative comparison of the two algorithms, vividly demonstrating the improvement of our algorithm in quality and speed from multiple viewpoints.

\begin{figure*}
    \centering
    \includegraphics[width=\textwidth]{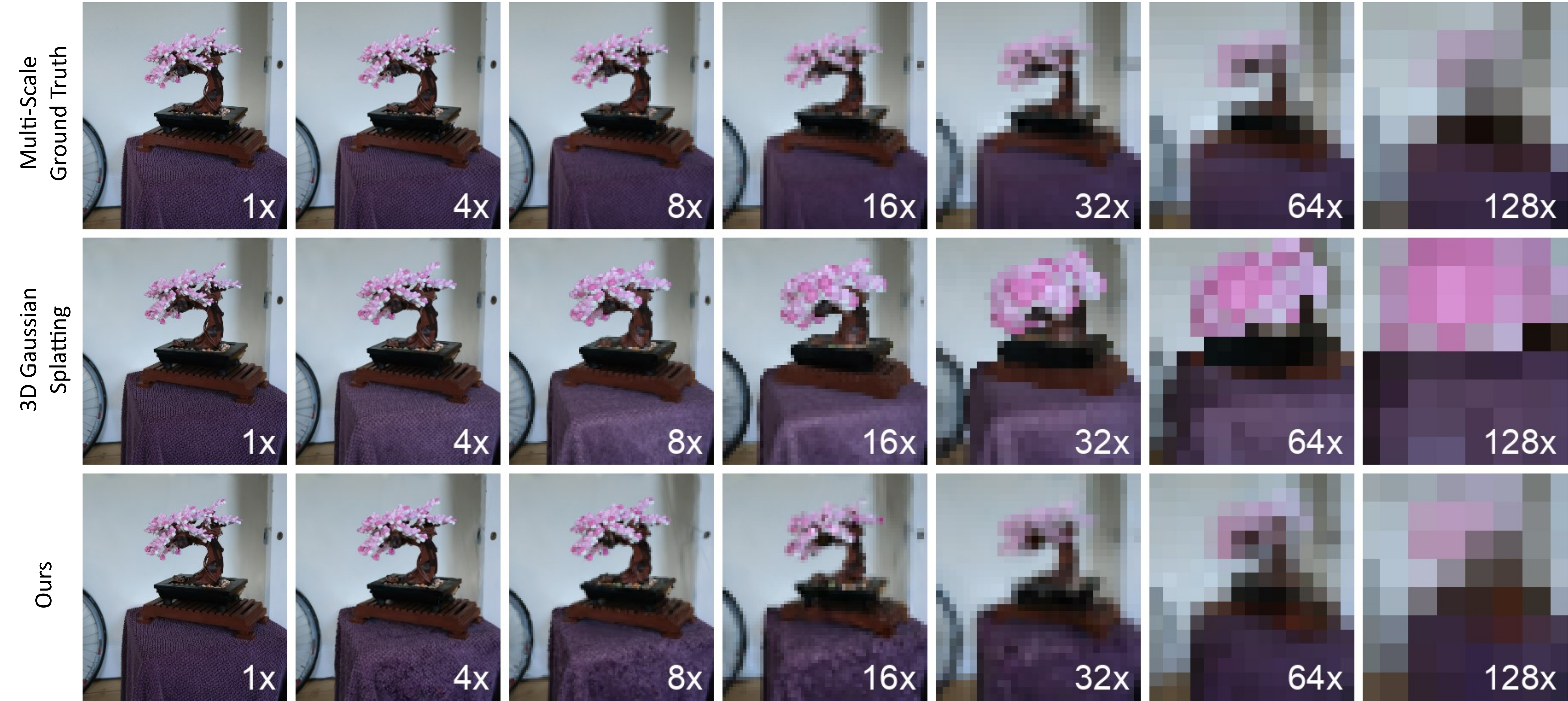}
    \vspace{-5mm}
    \caption{Qualitative Comparison on 360 dataset\cite{barron2022mipnerf360} for different resolution scales.}
    \label{fig:360_qualitative}
\end{figure*}

\begin{figure*}
    \centering
    \includegraphics[width=\textwidth]{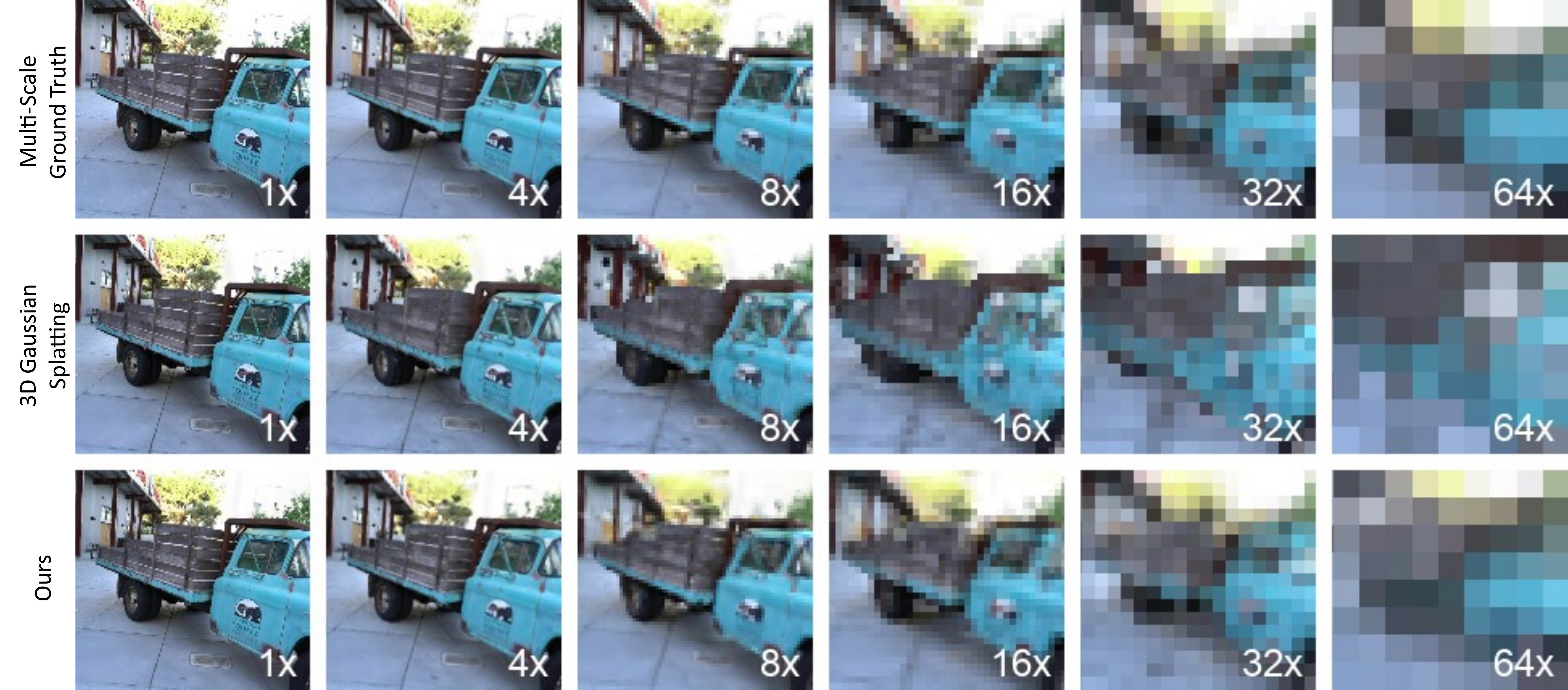}
    \vspace{-5mm}
    \caption{Qualitative Comparison on Tank and Temples dataset\cite{Knapitsch2017tank} for different resolution scales.}
    \label{fig:tnt_qualitative}
\end{figure*}

\begin{figure*}
    \centering
    \includegraphics[width=\textwidth]{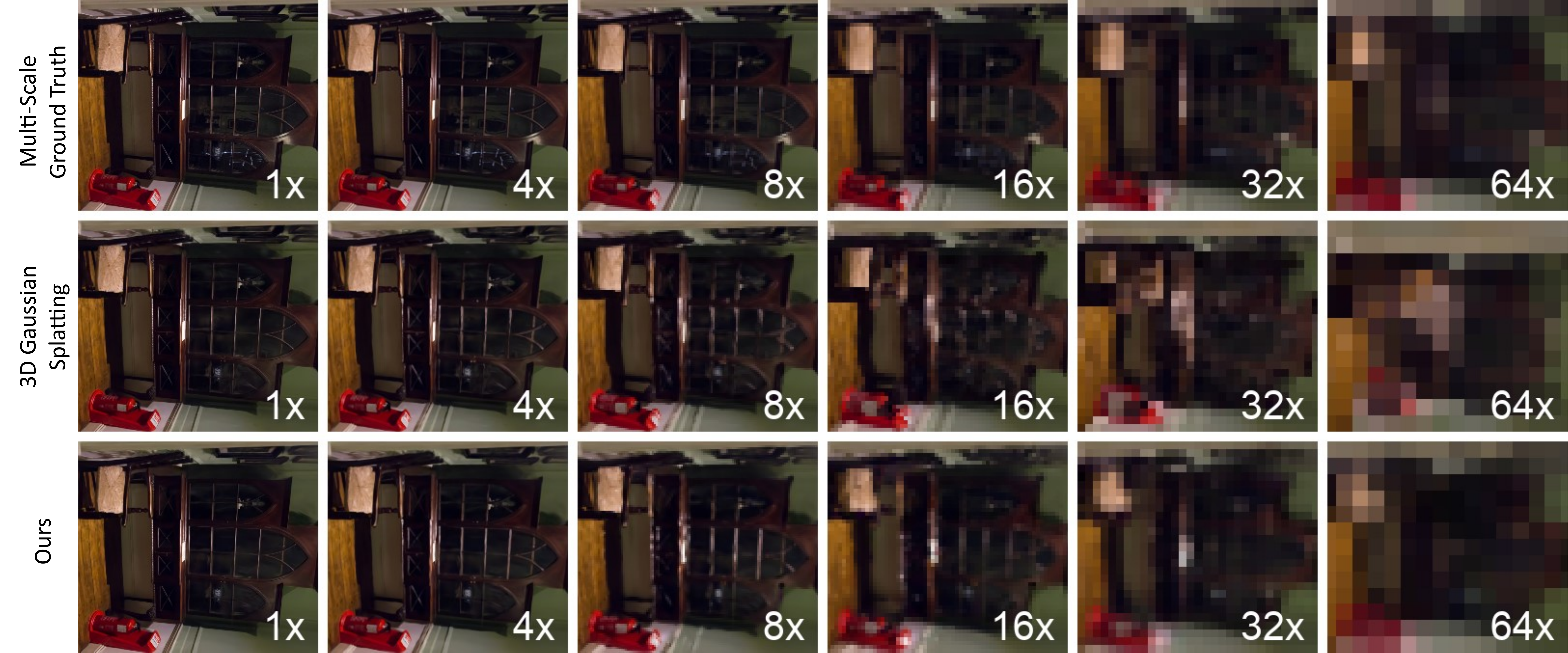}
    \vspace{-5mm}
    \caption{Qualitative Comparison on Deep Blending dataset\cite{deepblending} for different resolution scales.}
    \vspace{-5mm}
    \label{fig:db_qualitative}
\end{figure*}

\paragraph{Quantitative Comparison.}
As shown in \cref{tab:mipnerf-comparison}, \cref{tab:tnt-comparison}, and \cref{tab:db-comparison}, our method can achieve substantial quality and speed improvements compared to the original 3D Gaussian Splatting \cite{kerbl3Dgaussians} at lower resolutions. The quality and speed improvements become more pronounced as the resolution reduces, with the most noticeable 6-10dB PSNR and 20-30$\times$ speed gain at the 64$\times$ resolution scale. As the resolution reduces, the original splatting algorithm slows down while our method accelerates. %instead. 
The rendering quality and speed at the original resolution (1$\times$) remain comparable, indicating the effectiveness of our multi-scale Gaussians in representing both the high and low resolutions together.

\paragraph{Qualitative Comparison.}
We present the qualitative comparison with the original 3D Gaussian Splatting \cite{kerbl3Dgaussians} shown in \cref{fig:360_qualitative}, \cref{fig:tnt_qualitative}, and \cref{fig:db_qualitative}. At higher resolutions (1$\times$-8$\times$), both ours and the original algorithm can render the novel view rather faithfully. However, as the resolution reduces further(16$\times$-64$\times$), the original splatting algorithm produces severe artifacts, where the foreground becomes larger and larger, dominating the pixel colors as explained in \cref{sec:aliasing_cause}. In contrast, the images rendered by our method closely resemble the ground truth across all resolution scales. 

\paragraph{Ablations.}
To evaluate the effectiveness of the different components proposed, we present the ablation quantitative results in \cref{tab:mipnerf-comparison}, \cref{tab:tnt-comparison}, and \cref{tab:db-comparison} and the ablation qualitative results in the supplementary. The three ablation methods evaluated are named ``3DGS+MS Train", ``3DGS+Filter Small", and ``3DGS+Insert Large". ``3DGS+MS Train" reports the result with multi-scale training on top of the original 3D Gaussian splatting. The ``3DGS+Filter Small" reports the result with small Gaussian filtering using pixel coverage \textbf{and} the multi-scale training, which is needed to update the maximum and minimum pixel coverage. Similarly, the ``3DGS+Insert Large" reports the result with large Gaussian insertion \textbf{and} the multi-scale training. 

The ablation results reflect that %the
multi-scale training marginally improves %the
low-resolution rendering quality, but the rendering speed remains very slow. 
When filtering out the small Gaussians with multi-scale training, the speed at low resolution is increased by 20-30$\times$ with minimal rendering quality loss. The speed gain is caused by the considerably fewer Gaussians rendered.
When inserting the large Gaussians and training with multi-scale supervision, without the small Gaussian filtering, the rendering quality drops significantly because the details of the scene are covered with large Gaussians completely for all resolutions. However, when adding the large Gaussians together with the small Gaussian filtering, the rendering quality and speed at low resolution are enhanced significantly without jeopardizing the high-resolution quality. This indicates the effectiveness of all three components and the full method proposed. 
\section{Limitations}
\label{sec:limitations}
Since all Gaussian filtering of our proposed method relies on the pixel coverage, it can only be done after the initial splatting process when the coverage is calculated. Although the splatting of individual Gaussians are performed in parallel and does not takes more time at lower resolution, it is still a considerable overhead when rendering at a very low resolution. Even if a very small portion of the Gaussians are used for rendering in the end, all Gaussians still need to be splatted. This is the main reason why our rendering time is not decreased linearly as the resolution decreases. In our future work, we will look into a more lightweight criteria to filter small and large Gaussians before splatting them onto the screen to achieve an even faster rendering speed.

\vspace{-2mm}
\section{Conclusion}
\vspace{-1mm}
\label{sec:conclusion}
We analyzed the cause of the severe aliasing artifact and speed degradation of the existing 3D Gaussian splatting. We identified the key challenge of mitigating the aliasing for 3D Gaussian splatting lies in representing the scene with Gaussians of appropriate scale. Based on this observation, we propose to calculate the pixel coverage of 3D Gaussians during splatting and use this as a criteria for selective rendering. Gaussians that are too large or too small at the current rendering resolution are filtered for anti-aliasing and speed improvements. We also proposed to insert large Gaussians by aggregating small Gaussians during training to preserve the low frequency details and prevent part missing. Our experiments on various datasets support the effectiveness of our algorithm in rendering quality and speed at both high and low resolution, mitigating the severe aliasing artifact of the original 3D Gaussian splatting.

\paragraph{Acknowledgement.} This work is supported by the Agency for Science, Technology and Research (A*STAR) under its MTC Programmatic Funds (Grant No. M23L7b0021).
\maketitlesupplementary

\section{Video Comparison}
To better demonstrate the improvement of our algorithm in quality and speed for different resolutions, we include a video comparing our results with the original 3D Gaussian Splatting\cite{kerbl3Dgaussians} at multiple scenes from different views and resolutions. 

\section{Details of Gaussian Aggregation Algorithm}
Due to the space constraint of the main paper, some details of the Gaussian aggregation process are omitted. 
In this section, we will elaborate further with some examples to help the readers understand and reproduce our work.
The process consists of the following steps:
\paragraph{Render at Lower Resolution.}
Since we want to insert large Gaussians that are of appropriate size to be rendered at lower resolutions, we need to aggregate small Gaussians to form large Gaussians. Pixel coverage is used to determine whether a Gaussian is too small, we need to render all Gaussians first to calculate their pixel coverage at all training cameras.
For all coarse levels $l_m=[2,l_{max}]$, we render all Gaussians from $[1,l_m-1]$ at $4^{l_m-1}$ times downsampled resolution. 
For example, we render all Gaussians from level 1 to 3 at the $64\times$ downsampled resolution from all training cameras to add large Gaussians for level 4.
A Gaussian splatted to any of the training cameras with a pixel coverage $S_k$ smaller than $S_T$ is considered too small, and is included for the next step of aggregation.

\section{Theoretical Anti-aliasing Effectiveness of Gaussian Aggregation for 1D Signals}
\begin{figure}
    \centering
    \vspace{-5mm}
    \includegraphics[width=0.8\linewidth]{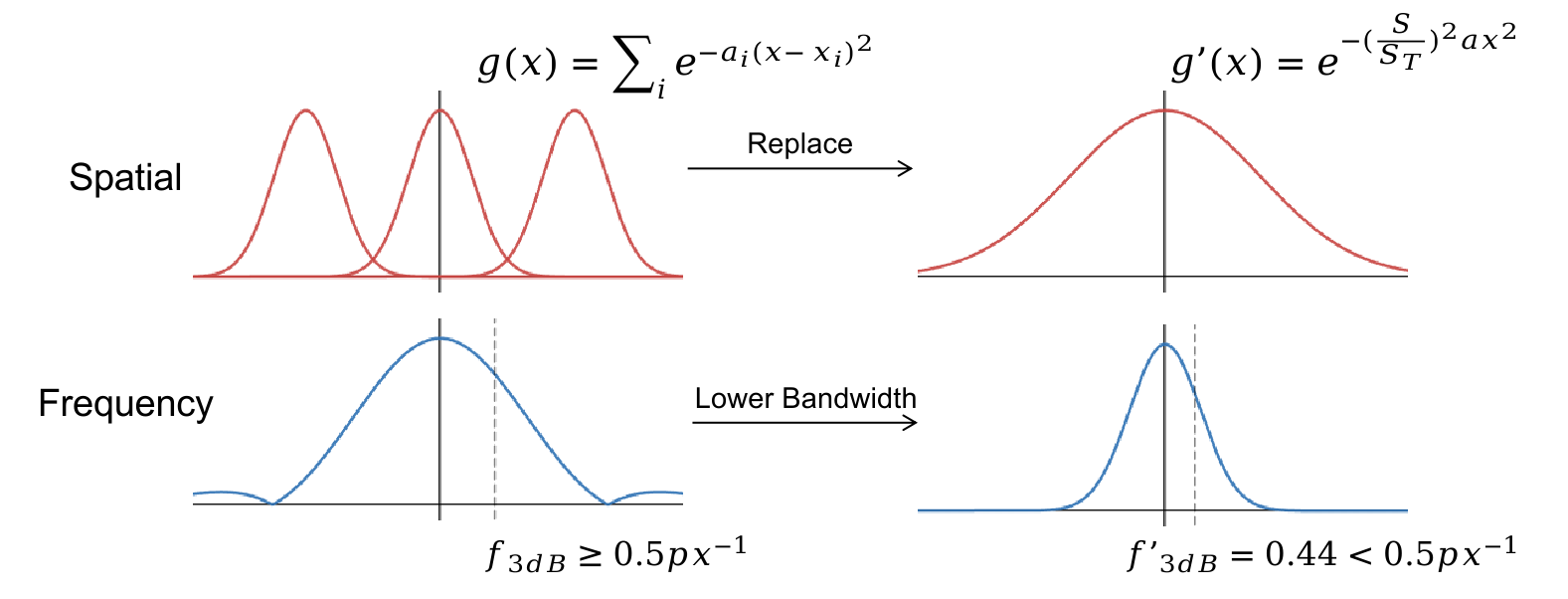}
    \vspace{-3mm}
    \caption{Instead of low-pass filters, we replace smaller Gaussians with fewer larger Gaussians, ensuring their bandwidth is below the aliasing threshold $0.5\mathrm{px}^{-1}$. }
    \vspace{-3mm}
    \label{fig:replace_gaussian}
\end{figure}

Our algorithm eschews low-pass filters for individual Gaussians as they do not mitigate the slow rendering speed. Instead, as shown in \cref{fig:replace_gaussian}, we opt to substitute smaller Gaussians with fewer, larger ones, reducing the signal bandwidth and the number of primitives rendered.
Heeding the reviewer's suggestion, we now delve deeper into the signal-processing analysis of our algorithm's anti-aliasing effect from first principles. Aliasing arises when a signal's bandwidth surpasses half the sampling frequency, as per Nyquist sampling theorem. 
Taking the mixture of 1D Gausssians $\Sigma_i e^{-a_i (x-x_i)^2}$ as an example, where $a_i=\frac{1}{2\sigma_i^2}$, we aim to prove that in our algorithm, they are consistently substituted with a Gaussian whose 3dB bandwidth is below the aliasing frequency threshold $0.5\mathrm{px}^{-1}$. 

According to our algorithm, the mixture of Gaussians is first aggregated into an average Gaussian $g_{avg}(x)=e^{-a(x-\mu)^2}$ with average $a=\frac{1}{N}\Sigma_i a_i$ and $\mu=\frac{1}{N}\Sigma_i x_i$. We can apply the Fourier transform to convert it to the frequency domain to become $\mathcal{F}[{g_{avg}}(x)](f)=\sqrt{\frac{\pi}{a}}e^{-\pi^2f^2/a}$.
% Taking a 1D Gaussian $g(x)=e^{-ax^2}$ as an example, where $a=\frac{1}{2\sigma ^2}$, we determine the commonly used 3dB bandwidth using the Fourier transform $\mathcal{F}[{g_s}(x)](f)=\sqrt{\frac{\pi}{a}}e^{-\pi^2f^2/a}$.
The 3dB bandwidth $f_{3dB}$ is the frequency where the magnitude is $1/\sqrt{2}$ of its peak magnitude. By solving
\vspace{-2mm}
\begin{equation}
    \sqrt{\frac{\pi}{a}}e^{-\pi^2f_{3dB}^2/a}=\frac{1}{\sqrt{2}}\sqrt{\frac{\pi}{a}},
    \vspace{-2mm}
\end{equation}
we find $f_{3dB}=\frac{1}{\pi}\sqrt{a\ln{\sqrt{2}}}$. 
% Aliasing occurs when $f_{3dB}\ge0.5f_{sample}=0.5\mathrm{px}^{-1}$.

Our algorithm then scales standard deviation up by $S_T/S$, where $S_T=2\mathrm{px}$ is the selective rendering threshold and $S$ is the pixel coverage of the Gaussian. We determine $S$ by calculating the size at its level set, solving $e^{-a(0.5S)^2}=1/N$ with $1/N=1/255$ on 8-bit color images. 
This yields $S=2\sqrt{\frac{1}{a}\ln{N}}$ and thus the scaled standard deviation becomes $\sigma '=\frac{2}{2\sqrt{\frac{1}{a}\ln{N}}}\sigma$. Given $a=\frac{1}{2\sigma^2}$, the scaled $a'=\frac{1}{a}\ln N \cdot a=\ln N$. 
Consequently, we calculate the 3dB bandwidth $f'_{3dB}$ of the scaled Gaussian as:
\vspace{-1mm}
\begin{equation}
    \begin{aligned}
        f'_{3dB}&=\frac{1}{\pi}\sqrt{a'\ln{\sqrt{2}}} =\frac{1}{\pi}\sqrt{\ln{N} \cdot \ln{\sqrt{2}}} \\
                &=0.441\mathrm{px}^{-1} < 0.5\mathrm{px}^{-1}.
    \end{aligned}
    \vspace{-1mm}
\end{equation}

This indicates that the bandwidth of the scaled Gaussians remains invariant to the attributes of the smaller Gaussians they replace, and is below half of the sampling frequency to avoid aliasing. While differing from the traditional low-pass filtering, our method is equally effective in anti-aliasing but more efficient in rendering.

\captionsetup[figure]{font=scriptsize}

\paragraph{Unbounded Scene Normalization.}
The Gaussians can be located at the range of $(-\infty, \infty)$ in unbounded scenes. This is not suitable for voxelization later as only a limited amount of voxels can be used. To normalize the unbounded space, the center region and the outer region are handled in different manners. The space bounded by a axis-aligned cube of length $B$ defined by the span of all training cameras is considered the center region, and the rest is considered the outer region. To preserve the structure in the center region, the coordinates are linearly scaled from $[-B,B]$ to $[-1,1]$. To normalize the unbounded outer region, the coordinates are non-linearly scaled from $(-\infty, \infty)$ to $(-2, 2)$. The exact normalization is as follows:
\begin{align}
    \mathbf{x}_{norm}=
    \begin{cases}
        \mathbf{x}/B,   & \text{if} \, max(|\mathbf{x}|) \le B \\
        2 - B/\mathbf{x},    & \text{otherwise}
    \end{cases}.
\end{align}

\paragraph{Voxelization.}
After the Gaussian positions are normalized to $[-2, 2]$, they need to be voxelized so that all Gaussians in one voxel are grouped together for the aggregation later.
The size of the voxel increases as the resolutions decrease because coarser levels require fewer larger Gaussians.
Specifically, when inserting large Gaussians for level $l_m$, the voxel size is chosen to be an empirical value of $(400/l_m)^3$. All Gaussians with their center in one voxel are grouped together for the next step. Although it is possible for a Gaussian to extent beyond the voxel while its center resides in the voxel, it is unlikely to reach too far as large Gaussians are filtered out in the earlier procedure.

\paragraph{Average Pooling and Enlargement}
After the small Gaussians are grouped in individual voxels, their parameters are averaged to create the large Gaussian. Specifically, the large Gaussian takes the average position, rotation, spherical harmonics features, opacity and scaling.
However, a new Gaussian would be too small if it remains at this scaling.
Consequently, we calculate the average pixel coverage of all the aggregated small Gaussians $S_{avg}$ using their pixel coverage derived earlier. 
The scaling of the new Gaussian is then enlarged by $S_T/S_{avg}$ for its pixel coverage to be approximately $S_T$, which is suitable to be rendered at level $l_m$.
This average pooling is not perfect, but simple and effective enough to produce a reasonable initialization for the multi-scale training later.

\newcolumntype{C}[1]{>{\centering\arraybackslash}p{#1}} % for center alignment
\newcolumntype{R}[1]{>{\raggedleft\arraybackslash}p{#1}} % for right alignment

\section{Qualitative Ablation Study}
\begin{figure*}
    \centering
    \includegraphics[width=0.8\textwidth]{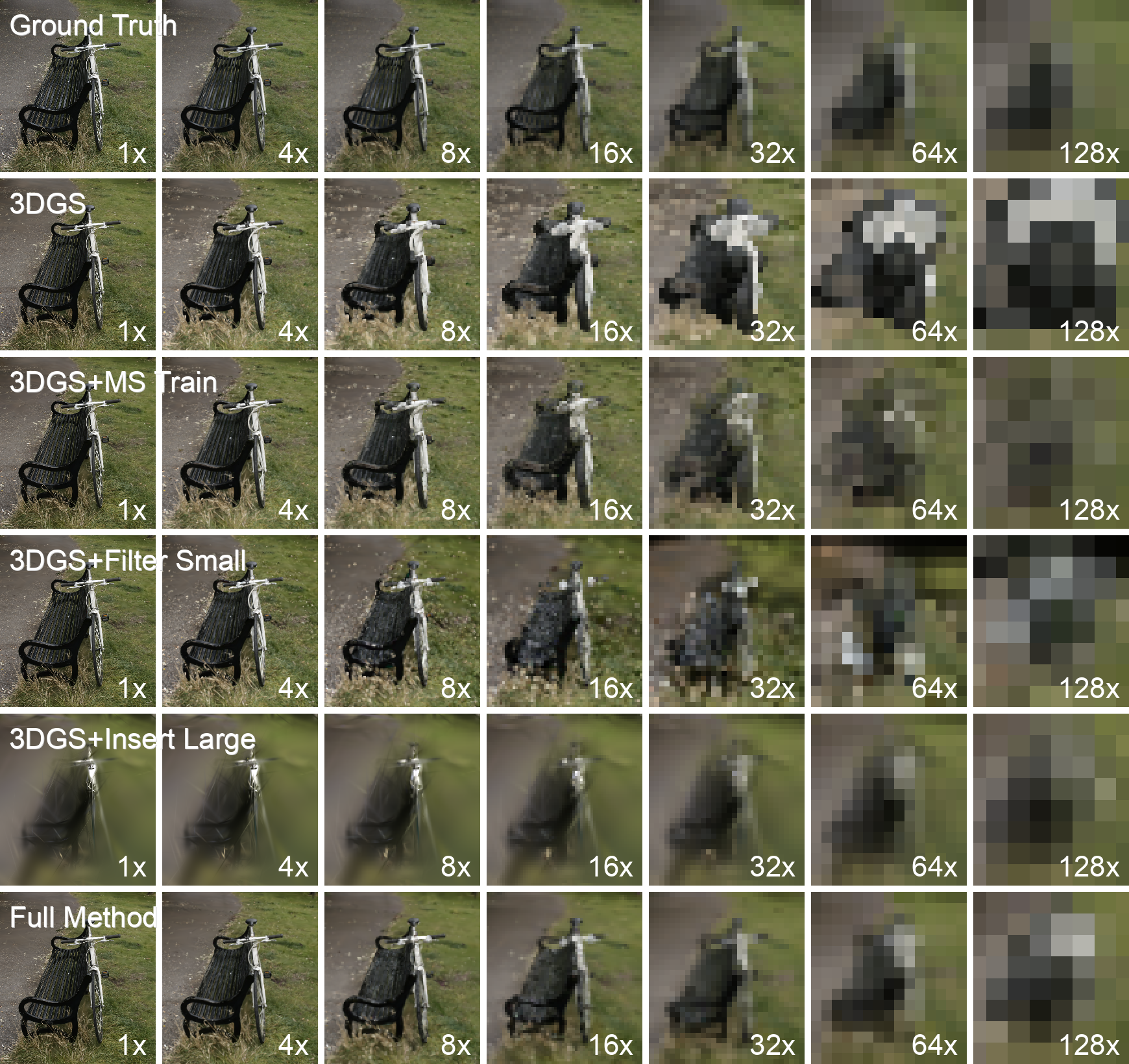}
    \caption{Qualitative ablation results of our proposed method on the "Bicycle" scene.}
    \label{fig:bicycle_ablation}
\end{figure*}

\begin{figure*}
    \centering
    \includegraphics[width=0.8\textwidth]{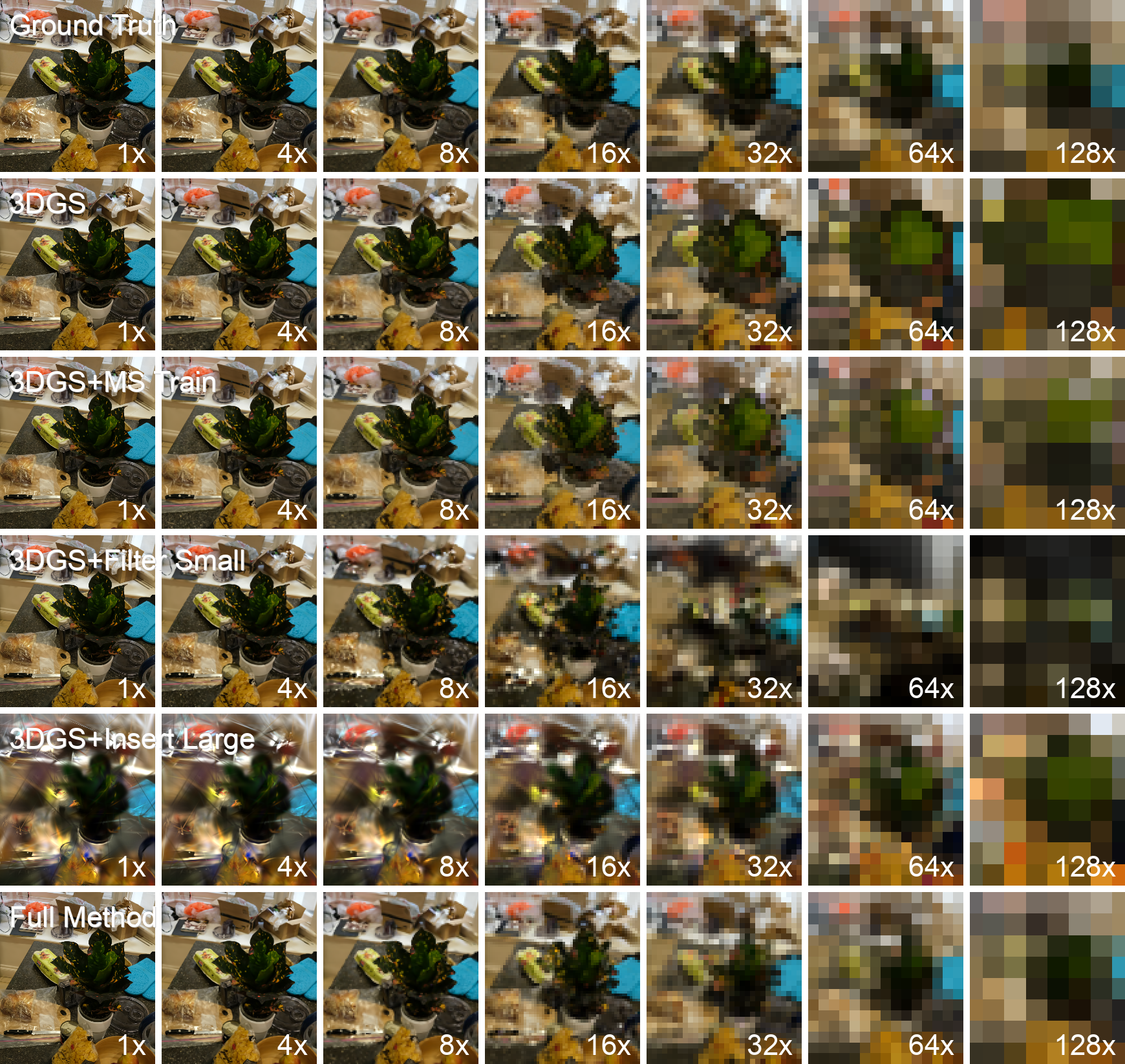}
    \caption{Qualitative ablation results of our proposed method on the "Counter" scene.}
    \label{fig:counter_ablation}
\end{figure*}

\begin{figure*}
    \centering
    \includegraphics[width=0.8\textwidth]{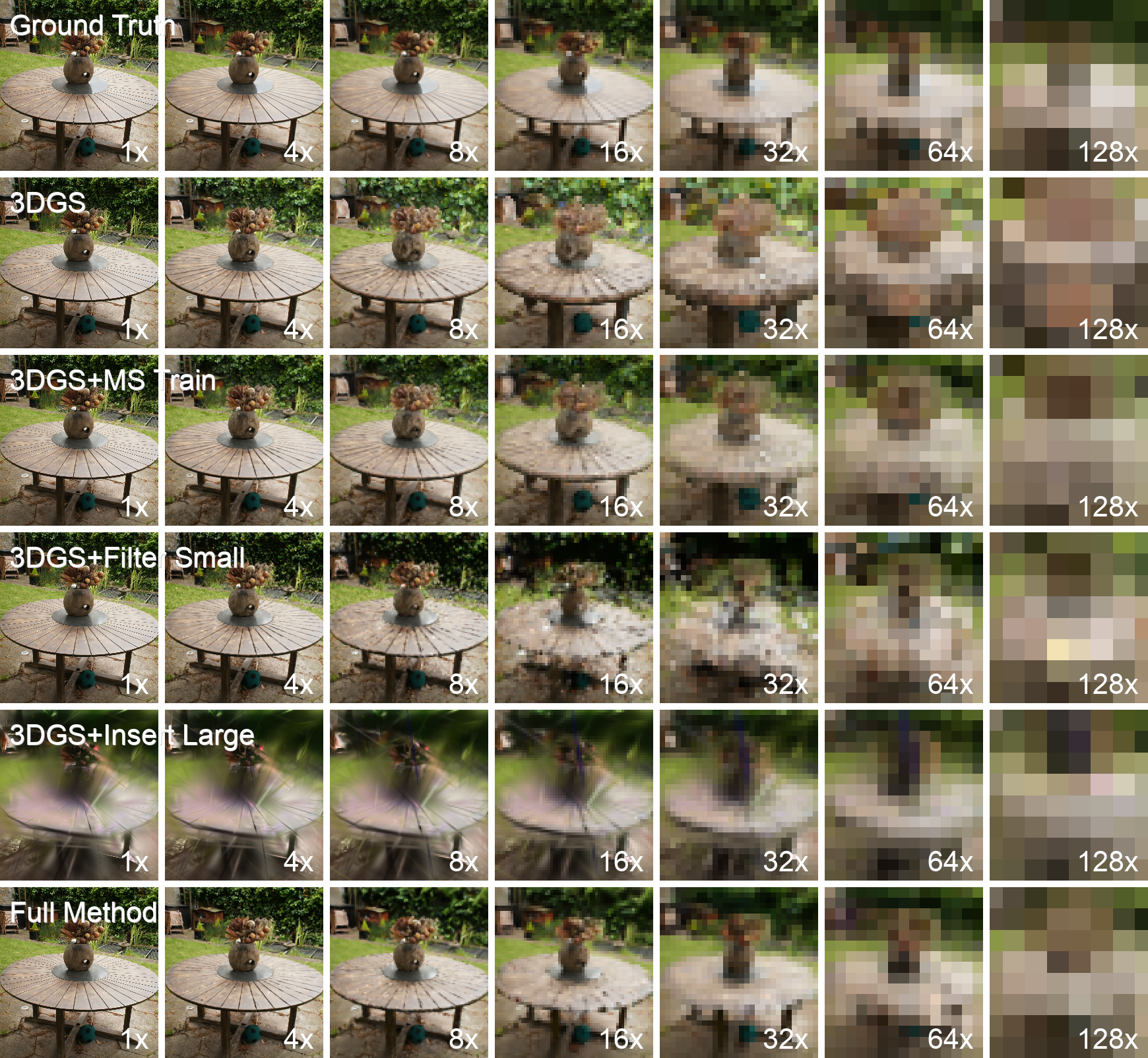}
    \caption{Qualitative ablation results of our proposed method on the "Garden" scene.}
    \label{fig:garden_ablation}
\end{figure*}

\begin{figure*}
    \centering
    \includegraphics[width=0.8\textwidth]{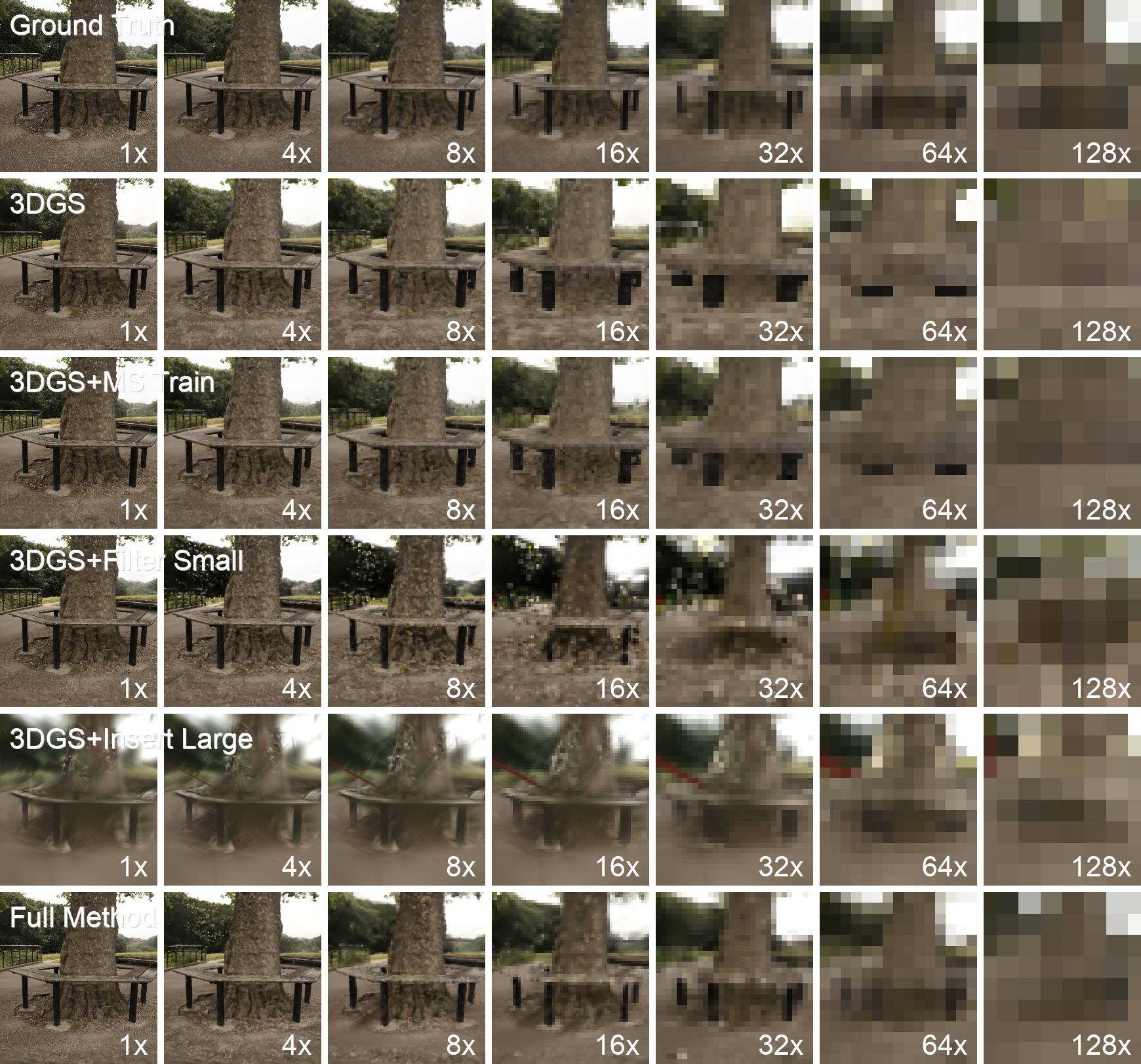}
    \caption{Qualitative ablation results of our proposed method on the "Treehill" scene.}
    \label{fig:treehill_ablation}
\end{figure*}

\begin{figure*}
    \centering
    \includegraphics[width=0.8\textwidth]{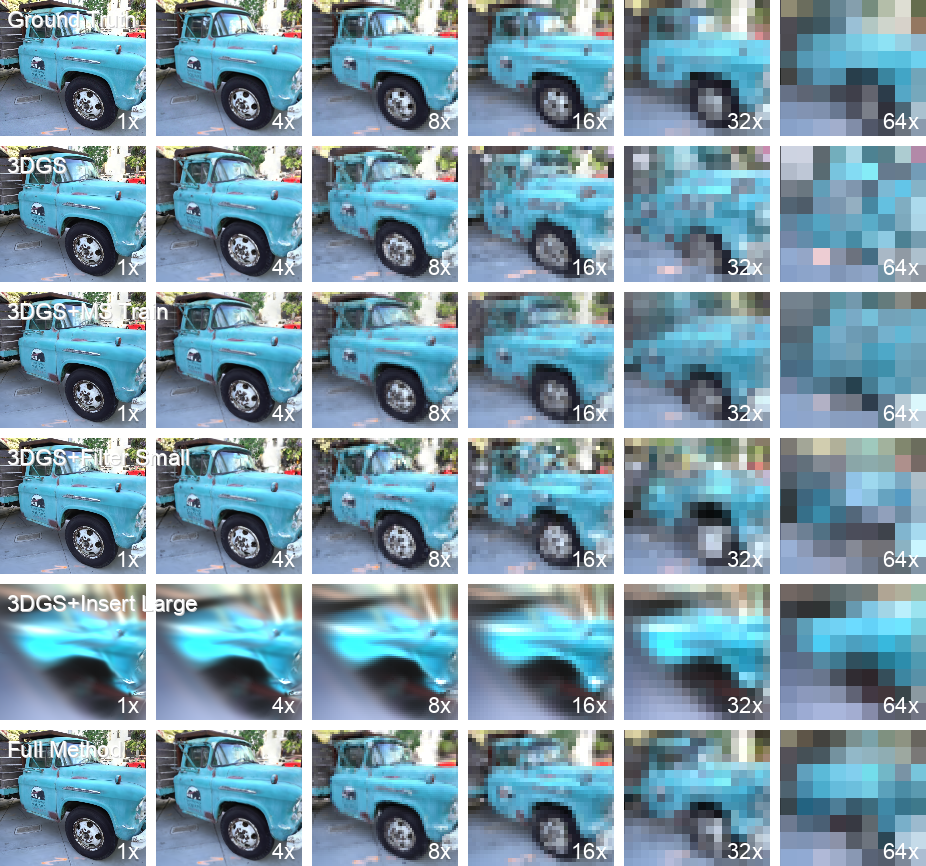}
    \caption{Qualitative ablation results of our proposed method on the "Truck" scene.}
    \label{fig:truck_ablation}
\end{figure*}

To better compare the effectiveness of each of our proposed module qualitatively, we present the rendering results of our method and various ablation models in~\cref{fig:bicycle_ablation}--\ref{fig:truck_ablation}. %, \cref{fig:counter_ablation}, \cref{fig:garden_ablation}, \cref{fig:treehill_ablation}, and \cref{fig:truck_ablation}. 
The ablation model design follows the experiment section in the main paper. Specifically, the ``+MS Train" model is trained using multi-scale images, but the Gaussians are only of a single scale as in 3D Gaussian Splatting \cite{kerbl3Dgaussians}. The low-resolution performance is slightly improved, but the rendering speed is as slow as the original method. The ``+Filter Small" model filters the small Gaussians based on the pixel coverage on top of the multi-scale training. It significantly accelerates the low-resolution rendering process, but the scene has some part missing as shown in the rendered images. The image rendered also has artifacts like black dots at low resolutions, caused by the filtered small Gaussians. The ``+Insert Large" model inserts the large Gaussians from aggregation on top of the multi-scale training. It has good rendering speed and quality at low resolutions, but the image rendered at high resolution is over-smoothed. This is caused by the finer level Gaussians not filtered out but optimized together with the inserted large Gaussians at low resolutions. Our "Full Method" overcomes the weakness of the ablation models and produces high-quality rendering at fast speed on both high and low resolutions. The small Gaussians filtered improves the speed, and the large Gaussians inserted improves the quality at low resolutions. The qualitative ablation supports the effectiveness of our proposed components.

\section{Quantitative Results on More Resolutions}
We present the quantitative results of our method, the original 3D Gaussian Splatting\cite{kerbl3Dgaussians}, and the various ablation methods on more downsampled resolutions. The resolutions include those that are not used during training which demonstrate the performance and robustness of our model. The experiments are conducted on MipNeRF-360 dataset \cite{barron2022mipnerf360} as shown in~\cref{tab:360_more_reso}, Tank and Temple dataset \cite{Knapitsch2017tank} as shown in~\cref{tab:tnt_more_reso}, and Deep Blending dataset \cite{deepblending} as shown in~\cref{tab:db_more_reso}.

\begin{table*}[!htp]
\centering
\caption{Quantitative comparison and ablation study on MipNeRF 360 dataset~\cite{barron2022mipnerf360} at more downsampled scales, with time in ``ms".}
\vspace{-3mm}
\label{tab:360_more_reso}
\begin{tabular}{lR{0.75cm}R{0.75cm}R{0.75cm}R{0.75cm}R{0.75cm}R{0.75cm}R{0.75cm}R{0.75cm}R{0.75cm}R{0.75cm}R{0.75cm}R{0.75cm}}
\toprule
Scale & \multicolumn{3}{c}{1x} & \multicolumn{3}{c}{2x} & \multicolumn{3}{c}{4x} & \multicolumn{3}{c}{8x} \\
\cmidrule(lr){2-4} \cmidrule(lr){5-7} \cmidrule(lr){8-10} \cmidrule(lr){11-13}
Metric & \small{PSNR$\uparrow$} & \small{LPIPS$\downarrow$} & \small{Time$\downarrow$} & \small{PSNR$\uparrow$} & \small{LPIPS$\downarrow$} & \small{Time$\downarrow$} & \small{PSNR$\uparrow$} & \small{LPIPS$\downarrow$} & \small{Time$\downarrow$} & \small{PSNR$\uparrow$} & \small{LPIPS$\downarrow$} & \small{Time$\downarrow$} \\
\midrule
3D Gaussian\cite{kerbl3Dgaussians} & \textbf{27.52} & \textbf{0.142} & 10.5 & 25.96 & \textbf{0.124} & 8.0 & 22.50 & 0.137 & 9.3 & 19.79 & 0.154 & 14.6 \\
3DGS + MS Train & 27.35 & 0.155 & 11.3 & 26.33 & 0.128 & 7.3 & 23.50 & \textbf{0.126} & 7.7 & 21.38 & 0.131 & 12.1 \\
3DGS + Filter Small & 27.40 & 0.153 & 10.0 & 26.42 & 0.129 & 6.8 & 23.81 & 0.149 & 5.4 & 21.73 & 0.175 & 5.1 \\
3DGS + Insert Large & 18.02 & 0.604 & 9.7 & 18.28 & 0.593 & \textbf{3.4} & 18.75 & 0.531 & \textbf{2.5} & 19.39 & 0.419 & \textbf{2.2} \\
Our Method & 27.39 & 0.155 & \textbf{9.1} & \textbf{26.44} & 0.134 & 6.3 & \textbf{24.82} & 0.132 & 5.4 & \textbf{24.44} & \textbf{0.112} & 5.1 \\
\bottomrule
\end{tabular}
\begin{tabular}{lR{0.75cm}R{0.75cm}R{0.75cm}R{0.75cm}R{0.75cm}R{0.75cm}R{0.75cm}R{0.75cm}R{0.75cm}R{0.75cm}R{0.75cm}R{0.75cm}}
\toprule
Scale & \multicolumn{3}{c}{16x} & \multicolumn{3}{c}{32x} & \multicolumn{3}{c}{64x} & \multicolumn{3}{c}{128x} \\
\cmidrule(lr){2-4} \cmidrule(lr){5-7} \cmidrule(lr){8-10} \cmidrule(lr){11-13}
Metric & \small{PSNR$\uparrow$} & \small{LPIPS$\downarrow$} & \small{Time$\downarrow$} & \small{PSNR$\uparrow$} & \small{LPIPS$\downarrow$} & \small{Time$\downarrow$} & \small{PSNR$\uparrow$} & \small{LPIPS$\downarrow$} & \small{Time$\downarrow$} & \small{PSNR$\uparrow$} & \small{LPIPS$\downarrow$} & \small{Time$\downarrow$} \\
\midrule
3D Gaussian\cite{kerbl3Dgaussians} & 17.79 & 0.149 & 27.9 & 16.30 & 0.084 & 55.2 & 15.23 & N.A. & 103.3 & 14.55 & N.A. & 123.2 \\
3DGS + MS Train & 20.21 & 0.115 & 22.8 & 19.80 & 0.060 & 45.6 & 19.38 & N.A. & 84.8 & 18.75 & N.A. & 100.1 \\
3DGS + Filter Small & 20.02 & 0.186 & 4.8 & 18.81 & 0.090 & \textbf{4.4} & 17.38 & N.A. & \textbf{4.6} & 16.13 & N.A. & \textbf{4.8} \\
3DGS + Insert Large & 20.23 & 0.256 & \textbf{2.7} & 21.17 & 0.081 & 4.6 & 21.53 & N.A. & 7.1 & 20.25 & N.A. & 9.4 \\
Our Method & \textbf{24.75} & \textbf{0.066} & 4.9 & \textbf{25.06} & \textbf{0.025} & 4.7 & \textbf{25.35} & N.A. & 4.9 & \textbf{22.55} & N.A. & 5.0 \\
\bottomrule
\end{tabular}
\end{table*}

\begin{table*}[!htp]
\centering
\caption{Quantitative comparison and ablation study on Tank and Temple dataset~\cite{Knapitsch2017tank} at more downsampled scales, with time in ``ms".}
\vspace{-3mm}
\label{tab:tnt_more_reso}
\begin{tabular}{lR{0.75cm}R{0.75cm}R{0.75cm}R{0.75cm}R{0.75cm}R{0.75cm}R{0.75cm}R{0.75cm}R{0.75cm}R{0.75cm}R{0.75cm}R{0.75cm}}
\toprule
Scale & \multicolumn{3}{c}{1x} & \multicolumn{3}{c}{2x} & \multicolumn{3}{c}{4x} & \multicolumn{3}{c}{8x} \\
\cmidrule(lr){2-4} \cmidrule(lr){5-7} \cmidrule(lr){8-10} \cmidrule(lr){11-13}
Metric & \small{PSNR$\uparrow$} & \small{LPIPS$\downarrow$} & \small{Time$\downarrow$} & \small{PSNR$\uparrow$} & \small{LPIPS$\downarrow$} & \small{Time$\downarrow$} & \small{PSNR$\uparrow$} & \small{LPIPS$\downarrow$} & \small{Time$\downarrow$} & \small{PSNR$\uparrow$} & \small{LPIPS$\downarrow$} & \small{Time$\downarrow$} \\
\midrule
3D Gaussian\cite{kerbl3Dgaussians} & 23.74 & \textbf{0.096} & 6.5 & 22.55 & 0.080 & 7.1 & 19.70 & 0.105 & 11.1 & 17.34 & 0.117 & 21.5 \\
3DGS + MS Train & 22.97 & 0.118 & 6.0 & \textbf{23.04} & 0.083 & 6.3 & 21.46 & \textbf{0.086} & 9.6 & 20.18 & \textbf{0.080} & 18.5 \\
3DGS + Filter Small & \textbf{23.78} & 0.100 & 5.6 & 22.76 & \textbf{0.079} & 5.1 & 20.12 & 0.107 & 4.5 & 18.62 & 0.122 & 4.4 \\
3DGS + Insert Large & 10.84 & 0.697 & \textbf{5.1} & 10.96 & 0.719 & \textbf{2.4} & 11.15 & 0.703 & \textbf{1.7} & 11.40 & 0.631 & \textbf{1.6} \\
Our Method & 23.46 & 0.111 & 7.6 & 22.44 & 0.095 & 5.6 & \textbf{21.92} & 0.087 & 4.7 & \textbf{20.88} & 0.082 & 4.6 \\
\bottomrule
\end{tabular}
\begin{tabular}{lR{0.75cm}R{0.75cm}R{0.75cm}R{0.75cm}R{0.75cm}R{0.75cm}R{0.75cm}R{0.75cm}R{0.75cm}}
\toprule
Scale & \multicolumn{3}{c}{16x} & \multicolumn{3}{c}{32x} & \multicolumn{3}{c}{64x} \\
\cmidrule(lr){2-4} \cmidrule(lr){5-7} \cmidrule(lr){8-10}
Metric & \small{PSNR$\uparrow$} & \small{LPIPS$\downarrow$} & \small{Time$\downarrow$} & \small{PSNR$\uparrow$} & \small{LPIPS$\downarrow$} & \small{Time$\downarrow$} & \small{PSNR$\uparrow$} & \small{LPIPS$\downarrow$} & \small{Time$\downarrow$} \\
\midrule
3D Gaussian\cite{kerbl3Dgaussians} & 15.61 & 0.068 & 43.4 & 14.45 & N.A. & 70.9 & 13.88 & N.A. & 82.6 \\
3DGS + MS Train & 18.56 & 0.049 & 37.4 & 17.41 & N.A. & 61.7 & 16.54 & N.A. & 71.7 \\
3DGS + Filter Small & 17.41 & 0.072 & 4.4 & 16.05 & N.A. & 4.5 & 14.95 & N.A. & 4.7 \\
3DGS + Insert Large & 11.73 & 0.447 & \textbf{1.7} & 12.14 & N.A. & \textbf{2.1} & 12.62 & N.A. & \textbf{2.5} \\
Our Method & \textbf{20.91} & \textbf{0.034} & 4.8 & \textbf{21.01} & N.A. & 5.4 & \textbf{19.67} & N.A. & 5.9 \\
\bottomrule
\end{tabular}
\end{table*}

\begin{table*}[!htp]
\centering
\caption{Quantitative comparison and ablation study on Deep Blending dataset~\cite{deepblending} at more downsampled scales, with time in ``ms".}
\vspace{-3mm}
\label{tab:db_more_reso}
\begin{tabular}{lR{0.75cm}R{0.75cm}R{0.75cm}R{0.75cm}R{0.75cm}R{0.75cm}R{0.75cm}R{0.75cm}R{0.75cm}R{0.75cm}R{0.75cm}R{0.75cm}}
\toprule
Scale & \multicolumn{3}{c}{1x} & \multicolumn{3}{c}{2x} & \multicolumn{3}{c}{4x} & \multicolumn{3}{c}{8x} \\
\cmidrule(lr){2-4} \cmidrule(lr){5-7} \cmidrule(lr){8-10} \cmidrule(lr){11-13}
Metric & \small{PSNR$\uparrow$} & \small{LPIPS$\downarrow$} & \small{Time$\downarrow$} & \small{PSNR$\uparrow$} & \small{LPIPS$\downarrow$} & \small{Time$\downarrow$} & \small{PSNR$\uparrow$} & \small{LPIPS$\downarrow$} & \small{Time$\downarrow$} & \small{PSNR$\uparrow$} & \small{LPIPS$\downarrow$} & \small{Time$\downarrow$} \\
\midrule
3D Gaussian\cite{kerbl3Dgaussians} & 29.65 & \textbf{0.094} & 8.6 & 29.41 & 0.065 & 6.6 & 27.48 & 0.066 & 7.5 & 24.67 & 0.076 & 11.3 \\
3DGS + MS Train & 29.46 & 0.102 & 6.6 & 29.42 & 0.069 & 4.8 & 28.18 & \textbf{0.062} & 5.3 & 26.15 & 0.065 & 8.0 \\
3DGS + Filter Small & 29.68 & 0.095 & 6.7 & 29.53 & \textbf{0.064} & 4.9 & 28.26 & 0.064 & 4.2 & 26.51 & 0.082 & 3.8 \\
3DGS + Insert Large & 20.59 & 0.379 & \textbf{4.6} & 20.67 & 0.381 & \textbf{2.2} & 20.83 & 0.336 & \textbf{1.6} & 21.07 & 0.263 & \textbf{1.7} \\
Our Method & \textbf{29.70} & 0.096 & 7.4 & \textbf{29.58} & 0.065 & 4.8 & \textbf{28.43} & 0.064 & 3.9 & \textbf{27.59} & \textbf{0.063} & 3.6 \\
\bottomrule
\end{tabular}
\begin{tabular}{lR{0.75cm}R{0.75cm}R{0.75cm}R{0.75cm}R{0.75cm}R{0.75cm}R{0.75cm}R{0.75cm}R{0.75cm}}
\toprule
Scale & \multicolumn{3}{c}{16x} & \multicolumn{3}{c}{32x} & \multicolumn{3}{c}{64x} \\
\cmidrule(lr){2-4} \cmidrule(lr){5-7} \cmidrule(lr){8-10}
Metric & \small{PSNR$\uparrow$} & \small{LPIPS$\downarrow$} & \small{Time$\downarrow$} & \small{PSNR$\uparrow$} & \small{LPIPS$\downarrow$} & \small{Time$\downarrow$} & \small{PSNR$\uparrow$} & \small{LPIPS$\downarrow$} & \small{Time$\downarrow$} \\
\midrule
3D Gaussian\cite{kerbl3Dgaussians} & 22.06 & 0.067 & 20.7 & 19.74 & N.A. & 36.3 & 17.75 & N.A. & 59.7 \\
3DGS + MS Train & 24.13 & 0.055 & 14.3 & 22.09 & N.A. & 24.8 & 20.03 & N.A. & 41.3 \\
3DGS + Filter Small & 24.52 & 0.078 & 3.6 & 22.01 & N.A. & 3.3 & 18.29 & N.A. & \textbf{3.2} \\
3DGS + Insert Large & 21.29 & 0.143 & \textbf{2.1} & 21.14 & N.A. & \textbf{2.8} & 20.10 & N.A. & 4.2 \\
Our Method & \textbf{27.66} & \textbf{0.036} & 3.4 & \textbf{27.22} & N.A. & 3.3 & \textbf{25.70} & N.A. & 3.4 \\
\bottomrule
\end{tabular}
\end{table*}

\section{Per-Scene Quantitative Results}
We present the per-scene decomposition of the quantitative results of our method and the original 3D Gaussian splatting \cite{kerbl3Dgaussians} in various resolutions. The experiments are carried on MipNeRF-360 dataset \cite{barron2022mipnerf360} as shown in~\cref{tab:360_per_scene}, Tank and Temple dataset \cite{Knapitsch2017tank} as shown in~\cref{tab:tnt_per_scene}, and Deep Blending dataset \cite{deepblending} as shown in~\cref{tab:db_per_scene}. The scenes chosen to be tested on follow the experiments carried out in the original 3D Gaussian splatting paper \cite{kerbl3Dgaussians}.

\begin{table*}[!htp]
\centering
\caption{Per-scene performance decomposition on MipNeRF-360 dataset\cite{barron2022mipnerf360}. Time measured in 'ms'.}
\vspace{-3mm}
\label{tab:360_per_scene}
\begin{tabular}{l|l|R{0.6cm}R{0.6cm}R{0.6cm}|R{0.6cm}R{0.6cm}R{0.6cm}|R{0.6cm}R{0.6cm}R{0.6cm}|R{0.6cm}R{0.6cm}R{0.6cm}|R{0.6cm}R{0.6cm}R{0.6cm}}
\toprule
 & Scale & \multicolumn{3}{c}{1x} & \multicolumn{3}{c}{4x} & \multicolumn{3}{c}{16x} & \multicolumn{3}{c}{64x} & \multicolumn{3}{c}{128x} \\
\cmidrule(lr){3-5} \cmidrule(lr){6-8} \cmidrule(lr){9-11} \cmidrule(lr){12-14} \cmidrule(lr){15-17}
Scene & Metric & \small{PSNR$\uparrow$} & \small{LPIPS$\downarrow$} & \small{Time$\downarrow$} & \small{PSNR$\uparrow$} & \small{LPIPS$\downarrow$} & \small{Time$\downarrow$} & \small{PSNR$\uparrow$} & \small{LPIPS$\downarrow$} & \small{Time$\downarrow$} & \small{PSNR$\uparrow$} & \small{LPIPS$\downarrow$} & \small{Time$\downarrow$} & \small{PSNR$\uparrow$} & \small{LPIPS$\downarrow$} & \small{Time$\downarrow$} \\
\midrule
garden & 3D-GS\cite{kerbl3Dgaussians} & \textbf{27.27} & \textbf{0.070} & 15.0 & 20.42 & 0.136 & 14.4 & 16.74 & 0.166 & 48.8 & 14.92 & N.A. & 200.9 & 14.29 & N.A. & 245.0 \\
garden & Ours & 27.16 & 0.080 & \textbf{11.8} & \textbf{23.99} & \textbf{0.112} & \textbf{7.8} & \textbf{26.41} & \textbf{0.044} & \textbf{7.5} & \textbf{24.79} & N.A. & \textbf{8.6} & \textbf{21.19} & N.A. & \textbf{9.6} \\
flowers & 3D-GS\cite{kerbl3Dgaussians} & \textbf{21.41} & \textbf{0.309} & 9.1 & 18.89 & 0.239 & 8.8 & 15.46 & 0.165 & 24.9 & 13.90 & N.A. & 93.2 & 13.69 & N.A. & 112.2 \\
flowers & Ours & 21.11 & 0.333 & \textbf{8.1} & \textbf{20.83} & \textbf{0.234} & \textbf{5.7} & \textbf{21.97} & \textbf{0.093} & \textbf{5.1} & \textbf{22.69} & N.A. & \textbf{4.9} & \textbf{21.82} & N.A. & \textbf{5.0} \\
treehill&3D-GS\cite{kerbl3Dgaussians}&22.60&\textbf{0.274}&10.0&21.63&\textbf{0.232}&9.7&18.71&0.193&24.6&16.19&N.A.&90.6&15.52&N.A.&97.0\\
treehill&Ours&\textbf{22.64}&0.291&\textbf{8.7}&\textbf{22.31}&0.239&\textbf{5.8}&\textbf{23.55}&\textbf{0.072}&\textbf{5.4}&\textbf{24.28}&N.A.&\textbf{4.9}&\textbf{22.27}&N.A.&\textbf{5.0}\\
bicycle&3D-GS\cite{kerbl3Dgaussians}&\textbf{25.15}&\textbf{0.164}&18.8&19.71&0.178&15.5&16.27&0.215&43.9&14.99&N.A.&163.8&15.15&N.A.&187.0\\
bicycle&Ours&24.44&0.210&\textbf{13.4}&\textbf{24.76}&\textbf{0.131}&\textbf{7.4}&\textbf{25.00}&\textbf{0.081}&\textbf{6.4}&\textbf{26.02}&N.A.&\textbf{6.5}&\textbf{21.56}&N.A.&\textbf{6.9}\\
counter&3D-GS\cite{kerbl3Dgaussians}&29.15&\textbf{0.099}&7.5&24.81&0.084&6.4&17.94&0.101&19.2&14.32&N.A.&60.4&13.39&N.A.&74.6\\
counter&Ours&\textbf{29.17}&0.100&\textbf{6.6}&\textbf{26.77}&\textbf{0.076}&\textbf{3.3}&\textbf{23.44}&\textbf{0.057}&\textbf{2.8}&\textbf{24.59}&N.A.&\textbf{2.7}&\textbf{21.14}&N.A.&\textbf{2.7}\\
kitchen&3D-GS\cite{kerbl3Dgaussians}&\textbf{31.70}&\textbf{0.064}&9.3&23.95&\textbf{0.081}&8.5&18.50&0.093&35.4&15.00&N.A.&124.4&14.15&N.A.&150.3\\
kitchen&Ours&31.64&0.064&\textbf{8.1}&\textbf{25.93}&0.089&\textbf{4.2}&\textbf{24.16}&\textbf{0.049}&\textbf{3.9}&\textbf{25.35}&N.A.&\textbf{3.3}&\textbf{21.50}&N.A.&\textbf{3.2}\\
room&3D-GS\cite{kerbl3Dgaussians}&\textbf{31.63}&\textbf{0.093}&8.0&26.60&0.057&5.1&19.50&0.096&12.0&15.50&N.A.&49.2&14.37&N.A.&70.8\\
room&Ours&31.51&0.094&\textbf{6.6}&\textbf{28.95}&\textbf{0.053}&\textbf{3.1}&\textbf{28.15}&\textbf{0.025}&\textbf{2.9}&\textbf{25.77}&N.A.&\textbf{2.9}&\textbf{21.82}&N.A.&\textbf{2.9}\\
stump&3D-GS\cite{kerbl3Dgaussians}&\textbf{26.75}&\textbf{0.138}&\textbf{10.6}&22.24&0.152&10.1&18.57&0.188&26.5&17.33&N.A.&95.2&16.97&N.A.&114.0\\
stump&Ours&26.59&0.152&12.9&\textbf{23.52}&\textbf{0.150}&\textbf{8.2}&\textbf{25.22}&\textbf{0.112}&\textbf{7.2}&\textbf{29.22}&N.A.&\textbf{7.1}&\textbf{29.09}&N.A.&\textbf{7.2}\\
bonsai&3D-GS\cite{kerbl3Dgaussians}&32.04&\textbf{0.065}&6.0&24.23&\textbf{0.075}&5.3&18.43&0.126&15.4&14.95&N.A.&52.4&13.46&N.A.&57.9\\
bonsai&Ours&\textbf{32.27}&0.067&\textbf{5.5}&\textbf{26.32}&0.106&\textbf{3.3}&\textbf{24.87}&\textbf{0.062}&\textbf{2.8}&\textbf{25.40}&N.A.&\textbf{2.9}&\textbf{22.53}&N.A.&\textbf{2.8}\\
\bottomrule
\end{tabular}
\end{table*}

\begin{table*}[!htp]
\centering
\caption{Per-scene performance decomposition on Tank and Temple dataset\cite{Knapitsch2017tank}. Time measured in 'ms'.}
\vspace{-3mm}
\label{tab:tnt_per_scene}
\begin{tabular}{l|l|R{0.6cm}R{0.6cm}R{0.6cm}|R{0.6cm}R{0.6cm}R{0.6cm}|R{0.6cm}R{0.6cm}R{0.6cm}|R{0.6cm}R{0.6cm}R{0.6cm}}
\toprule
 & Scale & \multicolumn{3}{c}{1x} & \multicolumn{3}{c}{4x} & \multicolumn{3}{c}{16x} & \multicolumn{3}{c}{64x} \\
\cmidrule(lr){3-5} \cmidrule(lr){6-8} \cmidrule(lr){9-11} \cmidrule(lr){12-14}
Scene & Metric & \small{PSNR$\uparrow$} & \small{LPIPS$\downarrow$} & \small{Time$\downarrow$} & \small{PSNR$\uparrow$} & \small{LPIPS$\downarrow$} & \small{Time$\downarrow$} & \small{PSNR$\uparrow$} & \small{LPIPS$\downarrow$} & \small{Time$\downarrow$} & \small{PSNR$\uparrow$} & \small{LPIPS$\downarrow$} & \small{Time$\downarrow$} \\
\midrule
truck & 3D-GS\cite{kerbl3Dgaussians} & \textbf{25.39} & \textbf{0.064} & \textbf{7.3} & 19.97 & 0.103 & 11.3 & 15.69 & 0.064 & 49.2 & 14.20 & N.A. & 89.1 \\
truck & Ours & 24.94 & 0.078 & 9.0 & \textbf{23.67} & \textbf{0.059} & \textbf{5.4} & \textbf{22.62} & \textbf{0.024} & \textbf{6.0} & \textbf{19.99} & N.A. & \textbf{8.6} \\
train & 3D-GS\cite{kerbl3Dgaussians} & \textbf{22.09} & \textbf{0.129} & \textbf{5.8} & 19.42 & \textbf{0.108} & 10.9 & 15.54 & 0.072 & 37.6 & 13.57 & N.A. & 76.1 \\
train & Ours & 21.98 & 0.144 & 6.2 & \textbf{20.17} & 0.114 & \textbf{3.9} & \textbf{19.21} & \textbf{0.044} & \textbf{3.5} & \textbf{19.36} & N.A. & \textbf{3.3} \\
\bottomrule
\end{tabular}
\end{table*}

\begin{table*}[!htp]
\centering
\caption{Per-scene performance decomposition on Deep Blending dataset\cite{deepblending} Time measured in 'ms'.}
\vspace{-3mm}
\label{tab:db_per_scene}
\begin{tabular}{l|l|R{0.6cm}R{0.6cm}R{0.6cm}|R{0.6cm}R{0.6cm}R{0.6cm}|R{0.6cm}R{0.6cm}R{0.6cm}|R{0.6cm}R{0.6cm}R{0.6cm}}
\toprule
 & Scale & \multicolumn{3}{c}{1x} & \multicolumn{3}{c}{4x} & \multicolumn{3}{c}{16x} & \multicolumn{3}{c}{64x} \\
\cmidrule(lr){3-5} \cmidrule(lr){6-8} \cmidrule(lr){9-11} \cmidrule(lr){12-14}
Scene & Metric & \small{PSNR$\uparrow$} & \small{LPIPS$\downarrow$} & \small{Time$\downarrow$} & \small{PSNR$\uparrow$} & \small{LPIPS$\downarrow$} & \small{Time$\downarrow$} & \small{PSNR$\uparrow$} & \small{LPIPS$\downarrow$} & \small{Time$\downarrow$} & \small{PSNR$\uparrow$} & \small{LPIPS$\downarrow$} & \small{Time$\downarrow$} \\
\midrule
drjohnson & 3D-GS\cite{kerbl3Dgaussians} & 29.14 & \textbf{0.106} & 10.1 & 27.23 & 0.079 & 9.3 & 22.73 & 0.078 & 26.3 & 18.60 & N.A. & 67.6 \\
drjohnson & Ours & \textbf{29.19} & 0.108 & \textbf{8.6} & \textbf{27.96} & \textbf{0.078} & \textbf{4.4} & \textbf{26.80} & \textbf{0.051} & \textbf{3.9} & \textbf{27.19} & N.A. & \textbf{3.8} \\
playroom & 3D-GS\cite{kerbl3Dgaussians} & 30.15 & \textbf{0.082} & 7.0 & 27.72 & 0.053 & 5.7 & 21.40 & 0.056 & 15.0 & 16.89 & N.A. & 51.8 \\
playroom & Ours & \textbf{30.20} & 0.084 & \textbf{6.2} & \textbf{28.89} & \textbf{0.051} & \textbf{3.4} & \textbf{28.53} & \textbf{0.020} & \textbf{3.0} & \textbf{24.22} & N.A. & \textbf{3.0} \\
\bottomrule
\end{tabular}
\end{table*}
\clearpage
\clearpage
{
    \small
    \bibliographystyle{ieeenat_fullname}
    \bibliography{main}
}


\end{document}